\pdfoutput=1

\documentclass[11pt]{article}

\usepackage[final]{acl}

\usepackage{times}
\usepackage{latexsym}

\usepackage[T1]{fontenc}

\usepackage[utf8]{inputenc}

\usepackage{microtype}

\usepackage{inconsolata}

\usepackage{graphicx}
\usepackage{booktabs} 
\usepackage{array}
\usepackage{pifont}
\usepackage{amssymb}
\usepackage[linesnumbered,ruled,vlined]{algorithm2e}
\usepackage{algpseudocode}
\usepackage{amsmath}
\usepackage{verbatim}
\usepackage{enumitem}
\usepackage{tcolorbox}
\usepackage{titlesec}
\usepackage{hyperref}
\usepackage{float}
\usepackage{tabularx}
\usepackage{subcaption}
\usepackage{soul}
\usepackage{longtable} 
\usepackage{booktabs}  
\usepackage{arydshln}
\usepackage{xcolor}

\title{\includegraphics[width=0.6cm]{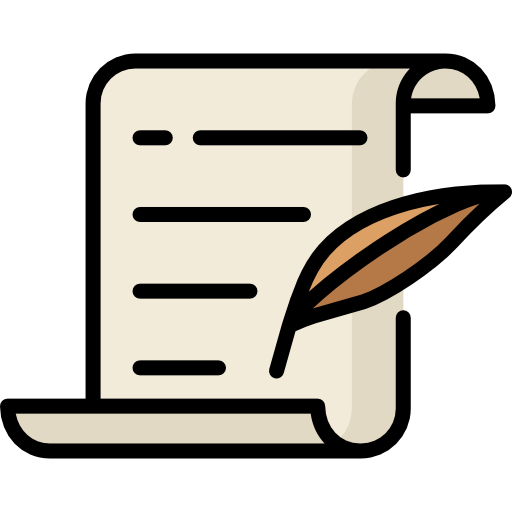} \textit{Whose story is it?} \\Personalizing story generation by inferring author styles}

\author{
    Nischal Ashok Kumar$^{1}$ \quad
    Chau Minh Pham$^{2}$ \quad
    Mohit Iyyer$^{1,2}$ \quad
    Andrew Lan$^{1}$ \\
    $^{1}$University of Massachusetts Amherst \quad
    $^{2}$University of Maryland, College Park \\
    \texttt{\{nashokkumar, andrewlan\}@cs.umass.edu}, 
    \texttt{\{chau, miyyer\}@umd.edu}
}

\newcommand{\dataname}{\emph{\textbf{Mythos}}}

\begin{document}
\maketitle

\begin{abstract} 
Personalization is critical for improving user experience in interactive writing and educational applications, yet remains understudied in story generation. We study the task of personalizing story generation, where our goal is to mimic an author's writing style, given other stories written by them. We collect \dataname, a dataset of 3.6k stories from 112 authors, with an average of 16 stories per author, across five distinct sources reflecting diverse story-writing settings. We propose a two-stage pipeline for personalized story generation: first, we infer authors' implicit writing characteristics and organize them into an \textit{Author Writing Sheet}, which is validated by humans to be of high quality; second, we simulate the author's persona using tailored persona descriptions and personalized story rules. We find that stories personalized using the Author Writing Sheet outperform a non-personalized baseline, achieving a 78\% win-rate in capturing authors' past style and 59\% in similarity to ground-truth author stories. Human evaluation supports these findings and further highlights trends, such as Reddit stories being easier to personalize, and the Creativity and Language Use aspects of stories being easier to personalize than the Plot.

\makebox[0pt][l]{\includegraphics[height=0.4cm]{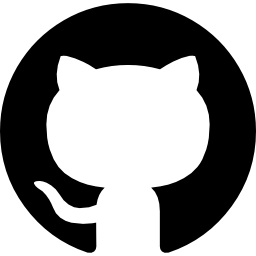}}
\hspace{1em} {\small \texttt{\href{https://github.com/Nish-19/Persona-Story-Gen}{github.com/Nish-19/Persona-Story-Gen}}}
\end{abstract} 

\section{Introduction}

\begin{figure*}[htbp]
\centering
\includegraphics[width=\linewidth]{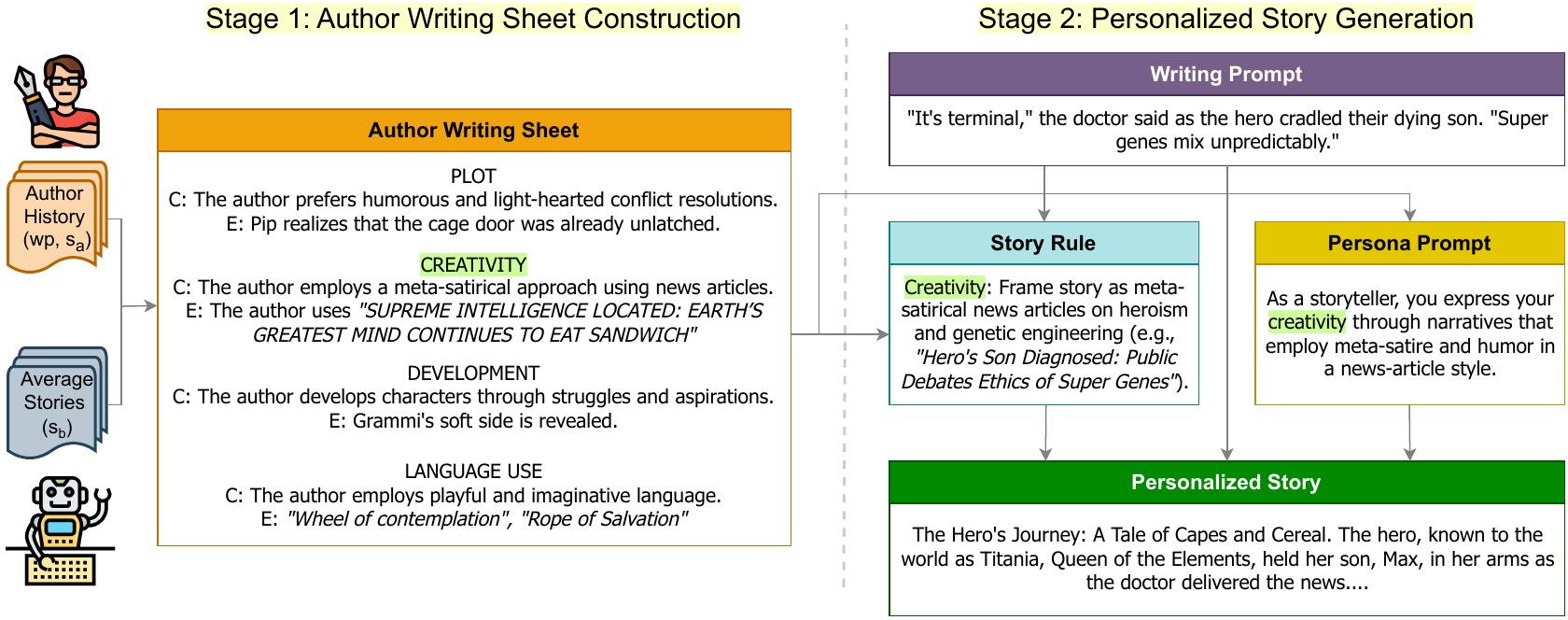}
\caption{Our two-stage pipeline for personalized story generation. Stage 1 constructs an Author Writing Sheet with Claim ($C$) and Evidence ($E$) pairs capturing the author’s story-writing characteristics across narrative categories. It is derived from the author’s history of writing prompts ($wp$), author-written stories ($s_a$), and LLM-generated Average Stories ($s_b$) representing a typical author's response to the same prompt. Stage 2 uses the Author Writing Sheet to role-play the author, incorporating tailored story rules and a persona description for personalized generation.} 
\label{fig:method}
\end{figure*}

Large Language Models (LLMs) are increasingly used in interactive systems to support human users in co-writing tasks \citep{fang2024sudowrite, yuan2022wordcraft, 10.1145/3491102.3501819}. A key challenge in these settings is to generate personalized outputs that align with individual user preferences \citep{yeh2024ghostwriter}, which improves engagement, agency, and writing quality \citep{10.1145/3706598.3713161}. Personalization is also key in educational applications to deliver effective feedback to students \citep{anderson1985intelligent, ashok-kumar-lan-2024-improving}, and may support users facing writer’s block \citep{yuan2022wordcraft} or help second language learners in learning a new language \citep{baffour-etal-2023-analyzing, su-etal-2023-reviewriter}. Despite these potentials, personalization in story writing has received limited attention.

Story-writing personalization faces several challenges, including a lack of existing resources for evaluating personalization and the inherent subjectivity in user preferences. To address these gaps, we construct \dataname, a dataset of 3.6k human-written stories ($\approx$1500 tokens each) from 112 authors (16 stories each\footnote{The reported average (16) is the mean of source-level stories-per-author counts (Table~\ref{tab:dataset-stats}), while the global average across all authors is 32 because Reddit has more stories per author than New Yorker and Narrative Magazine.}) across five distinct writing settings: Reddit, AO3, Storium, Narrative Magazine, and New Yorker. We split each author's stories chronologically into two sets: an earlier \emph{Profiling} set used to infer their writing characteristics, and a later \emph{Generation} set used to evaluate how well LLMs can emulate their distinctive writing behavior \citep{salemi-etal-2024-lamp}.

We systematically explore personalized story generation using LLMs with a two-stage pipeline as shown in Figure~\ref{fig:method}: First, we construct an \textit{Author Writing Sheet} to extract implicit writing characteristics from an author's past stories \citep{ramos-etal-2024-transparent}. The Author Writing Sheet covers four narrative dimensions: Plot, Creativity, Development, and Language Use \citep{pavis1998dictionary, huot2024agents}, and is organized as Claim-Evidence pairs grounded in story excerpts to summarize author preferences. This representation, inspired by Common Core Standards \citep{national2010common} and writing education research \citep{li2023teach}, offers a cost-effective and interpretable way to model authors' writing preferences. 
Second, we guide LLMs to role-play authors using persona prompts and story rules derived from the Author Writing Sheet.

We evaluate generated personalized stories in two ways: First, we use automatic evaluation with LLM-as-a-judge to assess faithfulness to the author's writing history and similarity to ground truth author story, along with traditional text similarity metrics on lexical overlap, story diversity, and style similarity. Second, we conduct a human evaluation to assess the generated story's similarity to the author's actual writings. 

We find that stories generated using the Author Writing Sheet outperform a non-personalized baseline: LLMs prefer them 78\% of the time in faithfulness and 59\% in similarity to ground truth, while human annotators prefer them 56\% of the times for similarity to ground truth. Our method is especially effective for Reddit stories and in capturing Creativity and Language Use, with a smaller impact on Plot. In summary, our contributions are:
\begin{itemize}[noitemsep, topsep=0pt]
    \item \textbf{\dataname}: A dataset of 3.6k stories with multiple stories per author, supporting both author profiling and personalized story generation evaluation.
    \item \textbf{Author Writing Sheet}: A cost-efficient, structured method for capturing author characteristics, grounded in narrative theory and validated through human evaluation.
    \item \textbf{Personalization Evaluation}: An evaluation combining automated and human assessments, showing the benefits of Author Writing Sheets. 
\end{itemize}

\section{Dataset Collection and Description}

To facilitate our analysis of \emph{personalized} story generation, we assemble \dataname, a dataset consisting of 3.6k stories written by 112 authors. To the best of our knowledge, \dataname~is the first dataset that identifies and connects multiple stories written by the same author (see Table~\ref{tab:compare-datasets} for comparison). 

\paragraph{Data Sources:} We include five diverse story-writing sources:
\textbf{(1) Reddit}, featuring stories from r/WritingPrompts, a widely-used resource for amateur story-writing research \citep{fan-etal-2018-hierarchical}, forming the largest subset with 3.2k stories,    
\textbf{(2) AO3}, featuring fanfiction based on popular franchises like \textit{Harry Potter},  
\textbf{(3) Storium}, featuring collaborative stories \citep{akoury-etal-2020-storium}, 
\textbf{(4) N.Mag} (Narrative Magazine), featuring professional stories from authors like Barry Gifford, and
\textbf{(5) N.York} (New Yorker), featuring expert-level storytelling from authors like Haruki Murakami.

\paragraph{Preprocessing:} To ensure high-quality and diverse content, we impose constraints on story length (ranging from 500 to 1500 words), limit publication dates to post-November 2022, and exclude stories with explicit content. To standardize formatting, non-Reddit stories are supplemented with writing prompts generated by GPT-4o, which are manually reviewed for accuracy. Stories are split chronologically, with 70\% used for \emph{profiling} writing characteristics and 30\% for testing \emph{generation}. Details on our dataset can be found in Appendix~\ref{app:dataset}.

\paragraph{Stories per Author Across Sources:} As summarized in Table~\ref{tab:dataset-stats}, Reddit authors contribute substantially more stories on average (54 per author), while expert sources such as N.Mag and N.York have only 3–4 stories per author due to limited public availability, reflecting natural differences in data accessibility between amateur and expert sources.

\begin{table*}[htbp]
\centering
\small
\begin{tabular}{m{4.95cm}m{0.50cm}m{4.95cm}m{0.75cm}m{0.75cm}m{0.75cm}}
\toprule
\textbf{Dataset Name} & \textbf{Size} & \textbf{Sources} & \textbf{Prompt Length} & \textbf{Story Length} & \textbf{Author IDs} \\
\midrule
WritingPrompts \citep{fan-etal-2018-hierarchical}       & 300k         & Reddit & 28             & 735              & \ding{55} \\ 
TELL ME A STORY \citep{huot2024agents}      & 230         & Writing Workshop      & 113             & 1498            & \ding{55}  \\ 
MirrorStories \citep{yunusov-etal-2024-mirrorstories}      & 1500         & Aesop’s fables      & 40             & 400            & \ding{55}  \\ 
Storium \citep{akoury-etal-2020-storium}      & 440k         & Storium online platform      & 247             & 247            & $\sim$  \\ 
\midrule
\dataname~(ours)  & 3.6k   & Reddit, AO3, Storium, N.Mag, N.York           & 50           & 1517          & \checkmark \\ 
\bottomrule
\end{tabular}
\caption{Comparison of our dataset with existing story-writing datasets. \ding{55} indicates no Author IDs associated with stories, and $\sim$ denotes having Author IDs but not having explicit links between them. \dataname~covers diverse story-writing settings and links stories by the same author, enabling future research on personalized story generation.}
\label{tab:compare-datasets}
\end{table*}

\section{Author Writing Sheet}
\label{sec:author-writing-sheet}

Motivated by the connection between writing education and personalization \citep{li2023teach}, we propose a method to infer authors' implicit story-writing tendencies based on their past stories in the profiling set, denoted as $P$. We group these tendencies into four narrative categories: Plot, Creativity, Development, and Language Use inspired by narrative theory \citep{pavis1998dictionary, huot2024agents}, to capture fine-grained, multi-dimensional aspects of how authors approach story-writing: 
\begin{itemize}[noitemsep, topsep=0pt]
    \item \textbf{Plot}: Story structure, conflict introduction, prompt engagement, and resolution.
    \item \textbf{Creativity}: Genre blending, unconventional prompt reinterpretation, and unique elements.
    \item \textbf{Development}: Character depth, emotional arcs, and immersive settings.
    \item \textbf{Language Use}: Diction, style, rhetorical devices, pacing, and dialogue.
\end{itemize}

Inspired by the Common Core Standards in English Language Arts \citep{national2010common}, each characteristic is represented as a Claim-Evidence pair, where the Claim summarizes the author's style and the Evidence are supporting story excerpts. 

Stage 1 in Figure~\ref{fig:method} (and Algorithm~\ref{alg:author_writing_sheet} in the Appendix) illustrates our method for generating the Author Writing Sheet (example in Table~\ref{tab:author_writing_sheet_sample}), where we iteratively process each story in an author's profiling set. 

We construct the sheet by merging characteristics across stories that highlight how the author's style differs from that of a typical, average author. 
For each writing prompt ($wp_t$) and its corresponding author story ($s_{at}$), we use the LLM with two distinct prompts: one prompt (\(\text{LLM\textsubscript{avg}}\)) elicits an average story ($s_{bt}$) that reflects the typical style of an author on that writing source, relying solely on source-specific priors acquired during pre-training; another prompt (\(\text{LLM\textsubscript{sheet}}\)) asks the same LLM to compare the author story ($s_{at}$) with the average story ($s_{bt}$) and extract corresponding Claim-Evidence pairs across narrative categories. This yields an intermediate Author Writing Sheet $A_t'$ \citep{shashidhar-etal-2024-unsupervised, krishna-etal-2020-reformulating}.

To construct the final Author Writing Sheet $A_{|P|}$, we iteratively merge each intermediate sheet $A_t'$ with the aggregated sheet $A_{t-1}$, following a moving update approach \citep{chang2024booookscore}. To reduce redundancy and manage output length, we prompt the LLM with a prompt (\(\text{LLM\textsubscript{combine}}\)) to group equivalent Claims, select the best Evidence, and retain ungrouped Claims. We further limit each narrative category to 10 Claim-Evidence pairs prioritizing grouped Claims to fit within the LLM’s context window in subsequent iterations resulting in a total of 40 Claim-Evidence pairs per author. Each Evidence is tagged with a timestamp indicating its source story. 

Similar to \textit{Knowledge Tracing} in education \citep{liu-etal-2022-open, 2023.EDM-posters.42, scarlatos2025exploring}, our method processes each story sequentially. Doing so enables efficient updates to the Author Writing Sheet without having to re-process prior stories every time a new story is analyzed \citep{yeh2024ghostwriter}. Therefore, this method reduces cost and mitigates long-context limitations in LLMs \citep{li2024long}; see Appendix~\ref{app:cost-analysis} for a detailed cost analysis. 

\paragraph{Validating Author Writing Sheet Quality:}

We conduct a human evaluation with three Upwork annotators to assess the quality of the Claim-Evidence pairs. Annotators evaluate each pair on two criteria: (1) whether the Claim can be reasonably inferred from the story (Yes/No), and (2) whether the Evidence supports the Claim (Yes/Partially/No). The study spans 42 authors across five sources, with one sampled story per author, yielding 188 annotated claims. To measure inter-rater reliability, 12 stories are annotated by all annotators. 
Results show unanimous agreement on the `Yes' label of Claims and moderate agreement on Evidence support (Krippendorff’s $\alpha = 0.57$), despite the task's subjectivity. Overall, 93\% of Evidence statements fully support their Claims, and the remaining 7\% provide partial support. These results indicate \emph{high precision} in constructing Claim-Evidence pairs and confirm the reliability and quality of the Author Writing Sheet; 
see Appendix~\ref{app:human-author-sheets} for more details including experimental design and interface (Figure~\ref{fig:labelstuido-sheet}), pilot studies, and results.


\section{Personalized Story Generation}
\label{sec:methods}

We now detail our proposed method for personalized story generation using the Author Writing Sheet, illustrated in Stage 2 of Figure~\ref{fig:method}. To generate personalized stories, we prompt an LLM (\(\text{LLM\textsubscript{story}}\)) with the writing prompt, story length, source-specific metadata (e.g., fanfiction for AO3), story rules as actionable instructions in direct second-person form categorized into four narrative categories (example in Table~\ref{tab:story_rules_sample}), and persona descriptions obtained from the Author Writing Sheet (example in Table~\ref{tab:persona_description_sample}). 
We experiment with four types of methods: 
(1) A non-personalized baseline, \emph{Average Author}; 
(2) Personalization baselines, \emph{RAG} and \emph{Delta}, that do not use the Author Writing Sheet; and 
(3) Our proposed methods, \emph{Sheet} and its variant \emph{Summary}. 
(4) An \emph{Oracle} method that loosely resembles the upper bound performance; See Appendix~\ref{app:story-gen} for the prompts used.


\paragraph{Non-Personalized: Average Author}
\hypertarget{sec:avg-author}{}
The \emph{Average Author} method serves as a non-personalized baseline reflecting typical author behavior learned during LLM pre-training. For each source, we prompt an LLM with an Average Author description summarizing writing characteristics obtained through audited GPT-4o prompting,\footnote{We ask GPT-4o for typical writing characteristics for each source and manually verify them.} verified manually \citep{wang-etal-2024-rolellm}. Example, for AO3, the Average Author is defined as ``respecting fandom tone and style while experimenting with tropes, unconventional pairings, and alternate universes.''

\paragraph{Personalization Baselines: RAG and Delta}

\paragraph{RAG}
\hypertarget{sec:rag}{}
The Retrieval-Augmented Generation (\emph{RAG}) baseline \citep{salemi-etal-2024-lamp} first retrieves the most similar writing prompt and author story from the profiling set using BM25 \citep{robertson2009probabilistic}. Following \citep{wang-etal-2024-rolellm}, the retrieved pair is used as a one-shot demonstration to elicit role-playing behavior from the LLM, mimicking the retrieved example's style.

\paragraph{Delta}
\hypertarget{sec:delta}{}
The \emph{Delta} method generates personalized story rules by contrasting the \emph{Average Author} story (see \hyperlink{sec:avg-author}{Average Author}) with the corresponding author-written story for each writing prompt in the profiling set. Following \citep{shashidhar-etal-2024-unsupervised}, we then use \emph{all} writing prompts in the profiling set, along with their corresponding generated story rules that are actionable instructions in direct second-person form, as few-shot demonstrations for the LLM, to generate personalized story rules for a new prompt in the generation set.

\paragraph{Our Methods: Sheet and Summary}

\paragraph{Sheet}
\hypertarget{sec:writing-sheet}{}
Our \emph{Sheet} method uses the Author Writing Sheet (Section~\ref{sec:author-writing-sheet}) for personalization (Stage 2 of Figure~\ref{fig:method}). First, we prompt an LLM (\(\text{LLM\textsubscript{persona}}\)) to generate a persona description that summarizes the author's story-writing style as a second-person narrative, included in the system prompt \citep{wang-etal-2024-rolellm, jiang2024evaluating}. Second, we prompt an LLM (\(\text{LLM\textsubscript{rule}}\)) to generate personalized story rules from the Author Writing Sheet tailored to the writing prompt in the generation set, included as constraints in the user prompt \citep{pham-etal-2024-suri} for \(\text{LLM\textsubscript{story}}\). Additionally, following \citep{richardson2023integrating}, we include a one-shot demonstration using the approach described in \hyperlink{sec:rag}{RAG}. 

\paragraph{Summary (Summ)}
\hypertarget{sec:writing-summary}{}
As a variant of the Author Writing Sheet, the \emph{Summ} method leverages the LLM's long-context capabilities \citep{ding2023longnet} by packing all past stories from the profiling set as input in the same prompt to generate an \emph{Author Writing Summary} in the same format as the Author Writing Sheet. Similar to the Sheet method, persona descriptions and story rules are derived from the Author Writing Summary and used as constraints for story generation, along with a one-shot demonstration. 

\paragraph{Reference Method: Oracle}
\hypertarget{sec:oracle}{}

The \emph{Oracle} method establishes an upper bound on personalization performance by using story rules derived directly from the ground-truth author story for each writing prompt. These rules are obtained by contrasting the \hyperlink{sec:avg-author}{Average Author} with the ground-truth story (see \hyperlink{sec:delta}{Delta}). We also include a one-shot demonstration following the approach described in \hyperlink{sec:rag}{RAG}.

\section{Experiments}
We now detail our experimental/evaluation settings. 

\subsection{Implementation Details}
We prompt GPT-4o in a chain-of-thought manner \citep{wei2022chain} with temperature 0 and a maximum token limit of 4096 for all the methods described above (Section~\ref{sec:methods}) to generate story rules and persona descriptions. Only for story generation (\(\text{LLM\textsubscript{story}}\)) across all methods, including \emph{Average Author}, we evaluate three different models: GPT-4o\footnote{For Reddit, we only consider 5 stories per author for generation totaling 300 stories, due to API cost constraints.} \citep{bubeck2023sparks}, Llama 3.1 8B, and Llama 3.1 70B \citep{dubey2024llama}, using a temperature of 0.7 and top\_p of 0.95 \citep{wang-etal-2024-rolellm} (See Appendix~\ref{app:llama-results} for results using Llama models). Additionally, we implement an \emph{ablation} variant of our proposed method for Sheet and Summ that excludes the persona description from the system prompt called \emph{Sheet-nP} and \emph{Summ-nP}, respectively.

\subsection{Automated Evaluation}

We evaluate the performance of personalized story generation using two automated evaluation methods: 
First, we use LLM-as-a-judge, which is a scalable alternative to human evaluation for complex open-ended text generation tasks \citep{zheng2023judging} and calculate the \textbf{win-rate} of a story generation method against the non-personalized baseline, \emph{Average Author}, on two aspects: 
\begin{itemize}[noitemsep, topsep=0pt]
    \item \textbf{Faithfulness to Writing History}: We evaluate how well the generated story aligns with the author history by measuring win-rates using Claims for each narrative category from the \hyperlink{sec:writing-summary}{Author Writing Summary} as the reference \citep{wang2023automated, yunusov-etal-2024-mirrorstories}.
    \item \textbf{Similarity to Author Story}: We assess how closely the generated story matches the author's ground-truth story by measuring win-rates using the ground-truth author story as the reference for each narrative category \citep{lyu2024href, shashidhar-etal-2024-unsupervised}.
\end{itemize}

For both aspects, we prompt OpenAI o4-mini\footnote{\url{https://openai.com/index/introducing-o3-and-o4-mini/}} in a chain-of-thought manner \citep{wei2022chain} to assign scores (1–5) to both stories. We randomly shuffle the order of stories in comparison, 
to avoid biases. Scores are broken down into four narrative categories: Plot, Creativity, Development, and Language Use \citep{saha-etal-2024-branch}. For each category, the story with the higher score is declared the winner, and the overall winner is based on the highest total score across all categories; two stories having the same total scores result in a tie. See Appendix~\ref{sec:llm-judge-prompts} for the prompts. Additionally, we evaluate personalization using \emph{traditional metrics} \citep{xie-etal-2023-next}, including lexical overlap, story diversity, and stylistic similarity. See Appendix~\ref{app:trad-merics} for details.


\section{Results and Discussion}

We now present experimental results. Our automatic evaluation shows that the proposed methods are effective, particularly for certain sources and narrative categories.


\subsection{Faithfulness to Writing History}

Table~\ref{tab:faith-auth-history} shows results for faithfulness to writing history among GPT-4o-personalized stories. We observe the following: 

\begin{table}[tp]
\centering
\small
\setlength{\tabcolsep}{3pt}
\renewcommand{\arraystretch}{1.1}
\begin{tabular}{p{1.5cm} @{}p{1.0cm}@{} p{0.80cm}@{} p{1.2cm}@{} p{1.2cm}@{} p{1.2cm}@{} p{0.8cm}@{}}
\toprule
\textbf{Method} & Reddit & AO3 & Storium & N.Mag & N.York & \emph{All} \\
\midrule
RAG & 40 & 52 & 42 & 21 & 40 & 39 \\
Delta & 56 & 60 & 50 & 64 & 53 & 57 \\
\midrule
Sheet & 74 & 76 & 73 & 85 & 80 & \underline{78} \\
Sheet-nP & 74 & 80 & 60 & 85 & 86 & \underline{78} \\
Summ & 84 & 96 & 87 & 85 & 80 & \emph{87} \\
Summ-nP & 76 & 86 & 85 & 100 & 100 & \emph{90} \\
\bottomrule
\end{tabular}
\caption{Win-rates (\%) for Faithfulness to Writing History vs.\ Average Author baseline for GPT-4o. `Sheet' (underlined) performs close to the upper bound established by the `Summ' method and significantly outperforms other personalization methods.}
\label{tab:faith-auth-history}
\end{table}

\paragraph{Sheet and Summ achieve best faithfulness to writing history:}  

The high performance of Summ sets a performance upper bound, since it relies on the \emph{Author Writing Summary}, the reference for evaluation, to generate the story. Sheet, despite not being explicitly conditioned on the Author Writing Summary, generalizes well (only a 9\% gap in win rate compared to Summ), which shows that it is capable in faithfully reflecting the author's writing history. RAG performs the worst since it relies on a single in-context example to mimic the author's style. Delta significantly outperforms RAG (by 18\%) but significantly underperforms Sheet (by 21\%). This result suggests that explicitly summarizing authors' stylistic characteristics systematically in the form of a Sheet is better than relying on rules extracted from all past author stories. See Appendix~\ref{app:sheet-best-faith} and Table~\ref{tab:sheet_vs_delta_faithfulness} for an example.

\paragraph{Persona descriptions benefit amateur sources more than expert sources for Summ:}

We see that personas have little impact on Sheet, since the overall win-rates for Sheet and Sheet-nP are identical. This result is likely attributed to using the Author Writing Summary, rather than the Sheet, as the reference for faithfulness evaluation. In contrast, for Summ, adding persona descriptions improves win-rates over Summ-nP for amateur sources such as Reddit, AO3, and Storium, but leads to lower performance for expert sources like the N.Mag and N.York. This result is likely due to personas for expert authors failing to capture the nuanced and subtle characteristics of their writing styles, which are reflected in their Author Writing Summary. See Appendix \ref{app:persona-summ-faith} and Tables~\ref{tab:persona_summ_faith_reddit}, \ref{tab:persona_summ_faith_nyork} for examples.

\subsection{Similarity to Author Story}

\begin{table}[tp]
\centering
\small
\setlength{\tabcolsep}{4pt}
\renewcommand{\arraystretch}{1.1}
\begin{tabular}{p{1.5cm} @{}p{1.0cm}@{} p{0.80cm}@{} p{1.2cm}@{} p{1.2cm}@{} p{1.2cm}@{} p{0.8cm}@{}}
\toprule
\textbf{Method} & Reddit & AO3 & Storium & N.Mag & N.York & \emph{All} \\
\midrule
Oracle & 86 & 84 & 78 & 79 & 87 & 83 \\
\midrule
RAG & 38 & 56 & \textbf{50} & 43 & 27 & 43 \\
Delta & 41 & 60 & 48 & 43 & 33 & 45 \\
\midrule
Sheet & \textbf{62} & \textbf{66} & 40 & \textbf{71} & \textbf{54} & \textbf{59} \\
Sheet-nP & 47 & 50 & 33 & 64 & 33 & 45 \\
Summ & 45 & 50 & 38 & 50 & 40 & 45 \\
Summ-nP & 43 & 53 & 45 & 64 & 20 & 45 \\
\bottomrule
\end{tabular}
\caption{Win-rates (\%) for Similarity to Author Story vs.\ Average Author baseline for GPT-4o. Best methods per source are \textbf{bolded}.}
\label{tab:sim-auth-story}
\end{table}

Table~\ref{tab:sim-auth-story} shows results on similarity to the author story among GPT-4o-personalized stories. We observe the following:

\paragraph{Sheet outperforms all other methods:} 
The Sheet method achieves the highest overall win-rate of 59\%, outperforming Summ by 14\%, Delta by 14\%, and RAG by 16\%. The advantage is more pronounced on Reddit, by far the largest subset of our dataset, where Sheet outperforms Summ by 17\%, Delta by 21\%, and RAG by 24\%. These results suggest that relying on a single in-context example (RAG) is insufficient for capturing personalized writing styles, particularly when authors have more varied profiles, as in Reddit. In contrast, for sources like Storium, where authors have fewer stories and less consistent styles, RAG is more competitive. The Delta method, which uses only story rules, struggles to capture subtle, repeated writing patterns that are better modeled through the longer and more detailed Author Writing Sheet. Finally, Summ's approach of identifying writing characteristics from all profiling stories at once, without contrasting against Average Author stories, is limited by LLMs' long-context reasoning capabilities \citep{shashidhar-etal-2024-unsupervised}. See Appendix~\ref{app:sheet-outperforms-sim} and Table~\ref{tab:creativity_example_sheet_wins} for an example. 

\paragraph{Sheet still lags behind the upper-bound Oracle method:}
Although the Sheet method performs best, there remains a substantial overall gap (of 24\%) compared to the \hyperlink{sec:oracle}{Oracle} method, which highlights the challenging nature of personalization in story generation. This gap likely arises because the Sheet cannot capture \emph{all} of an author's writing characteristics within the profiling set, due to generation length constraints (4096 tokens). Moreover, the Sheet does not explicitly rank characteristics based on their relevance to a new writing prompt, which limits its ability to adaptively emphasize the most salient aspects during generation. See Appendix~\ref{app:sheet-underperform} and Table~\ref{tab:oracle_sheet_gap} for an example.


\paragraph{Persona descriptions benefit Sheet:}
We see that Sheet outperforms its `nP' variant overall by 14\%, with a particular gain of 15\% on Reddit, the largest subset of our data. Persona descriptions, provided in the system prompt, offer additional control for personalized story generation by summarizing the author's general writing style independently of the specific writing prompt, complementing the story rules used in the user prompt. See Appendix~\ref{app:persona-sim-story} and Table~\ref{tab:persona_sim_comparison} for an example. In contrast, persona descriptions have limited impact on the Summ method, likely because the Author Writing Summary is less effective than the Author Writing Sheet at capturing subtle author characteristics.

\paragraph{Creativity and Language Use are easier to personalize:}

\begin{table}[H]
\centering
\small
\setlength{\tabcolsep}{2pt}
\renewcommand{\arraystretch}{1}
\begin{tabular}{lcccc}
\toprule
\textbf{Method} & Plot & Creativity & Development & Language Use \\
\midrule
Oracle & 74 & \textbf{82} & 69 & \textbf{88} \\
\midrule
Delta & 44 & \textbf{55} & 43 & \textbf{57} \\
Sheet & 49 & \textbf{74} & 51 & \textbf{62} \\
Summ & 44 & \textbf{51} & 50 & \textbf{51} \\
\bottomrule
\end{tabular}
\caption{Win-rates (\%) across narrative categories for Similarity to Author Story, averaged across sources. Best numbers for each method are \textbf{bolded}.}
\label{tab:category_comparison}
\end{table}

\begin{figure}[htbp]
\centering
\includegraphics[width=\linewidth]{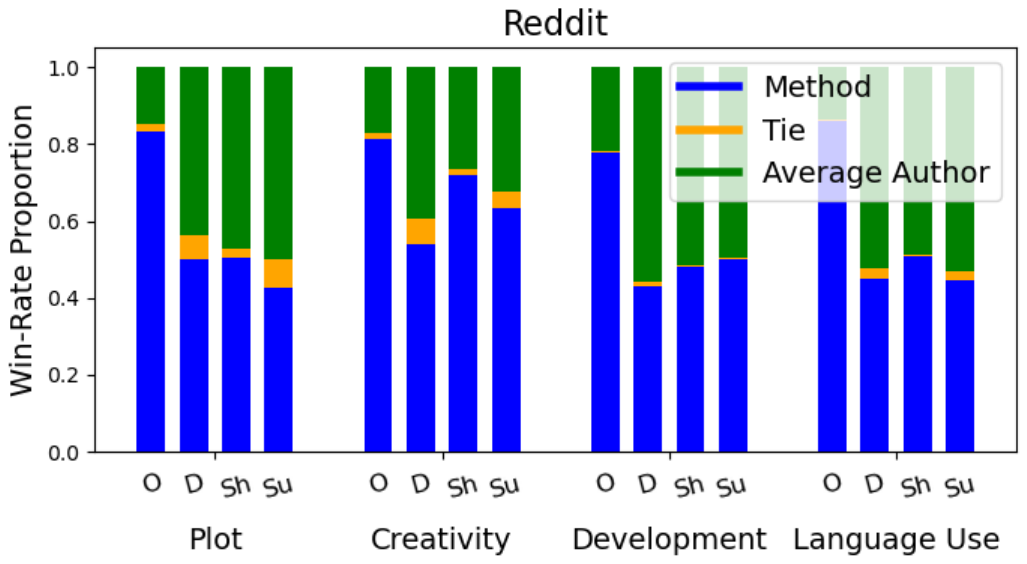}
\caption{Win-rates proportions across narrative categories for Similarity to Author Story for Reddit. O: Oracle, D: Delta, Sh: Sheet, Su: Summ.}\label{fig:reddit_win_rates}
\end{figure}

Table~\ref{tab:category_comparison} shows that all methods have higher respective win-rates on Creativity and Language Use than Plot and Development. This result likely arises because Creativity (e.g., unique elements introduced by the author) and Language Use (e.g., specific diction or use of dialogue) are less dependent on the writing prompt and thus more generalizable across stories \citep{huot2024agents}. In contrast, Plot (e.g., conflict introduction or resolution) and Development (e.g., character growth) are closely tied to the writing prompt, making generalization difficult, particularly when there is limited thematic overlap \citep{tian-etal-2024-large-language}.

Figure~\ref{fig:reddit_win_rates} shows that, for our largest story source, Reddit, the Sheet method achieves the highest win-rates across all categories except Development, where it is slightly outperformed by Summ. For Creativity, Sheet performs nearly as well as the Oracle, with only a 5\% win-rate gap between them. However, substantial gaps remain between Sheet and Oracle for other categories, which highlights the difficulty of the task and room for future improvement. See Appendix~\ref{app:cat-wise-results} and Figure~\ref{fig:win_rates_grid} for plots from other data sources.

\begin{table}[H]
\centering
\small
\setlength{\tabcolsep}{3pt}
\renewcommand{\arraystretch}{1}
\begin{tabular}{lccc}
\toprule
\textbf{Method} & \textbf{k = 1} & \textbf{k = 3} & \textbf{k = 5} \\
\midrule
Sheet & \textbf{70} & \textbf{67} & \textbf{66} \\
\midrule
RAG & 38 & 43 & 36 \\
Summ & 50 & 56 & 49 \\
\bottomrule
\end{tabular}
\caption{Effect of the number of few-shot demonstrations (\(k\)) for different methods, evaluated on 100 Reddit stories from 20 authors for Similarity to Author Story. The proposed Sheet method outperforms the best of other methods. Best results for each \(k\) are \textbf{bolded}.}
\label{tab:fewshot-results}
\end{table}

\paragraph{Sheet performs best across different numbers of few-shot demonstrations:}
We analyze the impact of the number of few-shot demonstrations (\(k\)) on model performance using 100 Reddit stories from 20 authors for different methods. As shown in Table~\ref{tab:fewshot-results}, the Sheet method achieves the highest win rate at \(k = 1\) (70\%), with a slight decline as \(k\) increases, consistent with observations by \citep{richardson2023integrating}. In contrast, both RAG and Summ peak at \(k = 3\) but drop at \(k = 5\), suggesting sensitivity to prompt length and context dilution. Even at their best settings, these methods lag behind Sheet by 27\% and 14\%, respectively, demonstrating the superiority of our proposed method.

\begin{table}[H]
\centering
\small
\setlength{\tabcolsep}{2pt}
\renewcommand{\arraystretch}{1}
\begin{tabular}{lccccc}
\toprule
\textbf{Method} & Plot & Creat. & Dev. & Lang. & Overall \\
\midrule
Sheet (full) & \textbf{48} & \textbf{74} & \textbf{45} & \textbf{60} & \textbf{70} \\
\midrule
no Plot & 37 & 65 & 45 & \textbf{61} & 49 \\
no Creat. & 46 & 63 & 51 & 52 & 53 \\
no Dev. & 37 & 57 & 36 & 46 & 39 \\
no Lang. & 38 & 61 & 45 & 46 & 44 \\
\bottomrule
\end{tabular}
\caption{Ablation study on the four narrative categories in the Author Writing Sheet, conducted on 100 stories from 20 Reddit authors for Similarity to Author Story. Each cell reports the win rate (\%) of the corresponding method (row) for each narrative category (column). \emph{Creat.}, \emph{Dev.}, and \emph{Lang.} denote \emph{Creativity}, \emph{Development}, and \emph{Language Use}, respectively. \emph{Overall} indicates the combined win rate across all categories. Method names such as ``No Plot'' indicate variants of the Sheet method with the corresponding category excluded. Best numbers for each category are \textbf{bolded}.}
\label{tab:ablation-categories}
\end{table}

\paragraph{All four narrative categories contribute to Sheet’s performance:}
To examine the contribution of each narrative category in the Sheet method for similarity to the author story, we conducted an ablation study on a subset of 100 stories from 20 Reddit authors. In this setting, we removed one category at a time from both the story rules and persona descriptions in our Sheet method while keeping the others fixed. As shown in Table~\ref{tab:ablation-categories}, we see that, the full model that uses all four categories achieves the highest overall win rate of 70\%, outperforming all ablated variants. We also see that, excluding any single category lowers performance on the corresponding narrative dimension, confirming that all four aspects, \textit{Plot}, \textit{Creativity}, \textit{Development}, and \textit{Language Use} jointly contribute to effective personalization. We further note that removing \textit{Development} yields the largest drop (31\% gap) compared to the full model, indicating that while \textit{Development} is harder to personalize, it is also essential, as its exclusion results in the lowest overall win rate and highlights its importance to effective personalization.

\subsection{Traditional Metrics}  
Table~\ref{tab:traditional-metrics} in the Appendix shows results with traditional metrics. Overall, we see that the lexical overlap and diversity scores 
are similar across methods, since all generations use the same LLM, resulting in overlapping lexical distributions that limit these metrics' sensitivity to stylistic differences \citep{zheng2023judging, xie-etal-2023-next}.

We also see that the Sheet achieves the \emph{best stylistic similarity} to both Author History and Author Story, measured by LUAR \citep{rivera-soto-etal-2021-learning}. This result validates the Sheet's ability to explicitly summarize stylistic deviations from an Average Author, leading to better personalization.

\section{Human Evaluation}
\label{sec:human-eval-story-gen}

\subsection{Setting}
We conduct a human evaluation on a subset of our data to assess narrative subtext and identify insights potentially missed by LLM judges \citep{chakrabarty2024art, subbiah-etal-2024-reading}. Annotators compare stories generated by personalization methods (\emph{Delta}, \emph{Sheet}, and \emph{Summ}) against the \emph{Average Author} method on similarity to the ground-truth author story \citep{lyu2024href}. For each comparison, annotators choose an overall winner by evaluating both stories on all four narrative categories, and provide justifications as free-form responses. Three annotators, recruited via Upwork, evaluated 45 author stories in total. There are 15 shared stories across annotators to assess agreement, and 10 were unique to each annotator to expand coverage \citep{song-etal-2024-veriscore}. Each author story appeared in three different pairs (one per personalization method), resulting in 135 annotated story pairs. See Appendix~\ref{app:human-eval-story-gen} for details on experiment design and the interface (Figure~\ref{fig:labelstuido-human-story}). 

\subsection{Results}

\begin{table}[tp]
\centering
\small
\setlength{\tabcolsep}{4pt}
\renewcommand{\arraystretch}{1.1}
\begin{tabular}{p{1.2cm} @{}p{1.0cm}@{} p{1.0cm}@{} p{1.2cm}@{} p{1.2cm}@{} p{1.2cm}@{} p{0.8cm}@{}}
\toprule
\textbf{Method} & Reddit & AO3 & Storium & N.Mag & N.York & \emph{All} \\
\midrule
Delta & 44-44 & 33-33 & \textbf{33}-33 & 33-33 & 67-11 & 42-31 \\
Sheet & \textbf{67}-33 & \textbf{44}-22 & 22-33 & \textbf{67}-33 & \textbf{78}-11 & \textbf{56}-27 \\
Summ & 44-33 & 33-44 & \textbf{33}-22 & 44-22 & 67-22 & 44-29 \\
\emph{Overall} & 52-37 & 37-33 & 30-30 & 48-30 & 70-15 & 47-29 \\
\bottomrule
\end{tabular}
\caption{Win-rates (\%) for Similarity between GPT-4o-generated and Author stories, evaluated by Humans. Each cell (`X-Y') shows the method's (`X') and Average Author's (`Y') win-rates, with ties as remainder. Best methods per source are \textbf{bolded}.}
\label{tab:human-eval}
\end{table}

Table~\ref{tab:human-eval} shows results from our human evaluation. Sheet is clearly the best method, outperforming Average Author by 29\% across all sources. On Reddit, our largest source, it reaches a 67\% win-rate, further confirming its advantage.


\paragraph{Fair inter-rater and model-human agreement:}
Fleiss’ Kappa among human annotators and Cohen's Kappa between human annotations and the LLM judge (o4-mini) both range from 0.2 to 0.4, indicating fair agreement \citep{landis1977measurement}. This result is likely due to the inherent subjectivity of evaluating long-form texts \citep{subbiah-etal-2024-storysumm}, where annotators prioritize different narrative aspects when choosing the overall winner. The scale of human evaluation, constrained by annotation costs (\$639), also contributes to variability. See Appendix \ref{sec:human-qual-examples} for detailed examples. 

\paragraph{Creativity, and Language use are easier to personalize:} 
Same as LLM-as-a-judge, human annotators also favor personalized stories for Creativity and Language Use, noting their stronger use of deeper symbolism, thematic richness, layered narratives, and expressive language. However, Average Author was preferred when personalized methods incorrectly introduce elements not present in the author's ground-truth story, suggesting deviations from the author's usual style. See Table~\ref{tab:annotator_story_comments_analysis} in the Appendix for detailed analysis of annotator free-form comments in each narrative category. 

\paragraph{Reddit and New Yorker are easier to personalize than other sources:}
Reddit and New Yorker show the highest overall win-rates for the personalized methods (52\% and 70\%, respectively). Reddit benefits from the broad thematic variety of its crowd-sourced writing prompts (see Figure~\ref{fig:reddit-themes} in the Appendix), which leads to diverse author writing styles and makes authors easier to distinguish from the Average Author (see Table~\ref{tab:exp-all-correct}). New Yorker, with stories written by experts, benefits from the use of advanced narrative devices such as subtext, which humans can evaluate more reliably than LLMs \citep{subbiah-etal-2024-reading} (see Table~\ref{tab:exp-all-diff}). 
%
AO3 and Storium show lower overall win-rates for personalized methods. For AO3, LLMs' familiarity with common fanfiction tropes\footnote{\url{https://archiveofourown.org/admin_posts/25888}} strengthens the Average Author baseline (see Table~\ref{tab:exp-all-wrong}). For Storium, the small profile size and weak stylistic differentiation among authors beyond open-ended endings \citep{xu2025echoes, tian-etal-2024-large-language} limit the effectiveness of personalization methods (see Table~\ref{tab:exp-all-tie}).
These findings suggest that personalization is most effective when authors explore diverse topics or display distinct traits, such as complex narratives, creativity, or advanced language use. Such traits, reflected in past writings, provide clearer signals for personalizing their style.

\section{Related Work}

\subsection{Personalization and Role-Playing}
Recent works have introduced benchmark datasets for personalizing LLM outputs in tasks like email and news writing, focusing on shorter outputs (e.g., 300 tokens for product reviews \citep{kumar2024longlamp}, 850 for news writing \citep{shashidhar-etal-2024-unsupervised}). These methods infer user traits from history for task-specific personalization \citep{sun-etal-2024-revealing, sun-etal-2025-persona, pal-etal-2025-beyond, li2023teach, salemi2025reasoning}. In contrast, we address the more subjective task of long-form story writing, with author stories averaging 1500 tokens. Unlike prior role-playing approaches that rely on predefined personas (e.g., Tony Stark, Confucius) \citep{wang-etal-2024-rolellm, sadeq-etal-2024-mitigating, tu2023characterchat, xu2023expertprompting}, we propose a novel method to infer story-writing personas from an author's history to guide role-playing.

\subsection{Story Understanding and Generation}  
Prior work on persona-aware story generation \citep{yunusov-etal-2024-mirrorstories, zhang-etal-2022-persona, chandu-etal-2019-way} defines personas using discrete attributes like personality traits, demographics, or hobbies. Similarly, \citep{zhu-etal-2023-storytrans} explore style transfer across predefined domains (e.g., fairy tales, martial arts, Shakespearean plays). In contrast, we mimic an individual author's style based on their history. Our approach differs by (1) inferring long-form author personas descriptions of style from past works, rather than relying on demographics, and (2) handling long-form story generation, averaging 1500 tokens per output, beyond typical lengths in prior work.

\section{Conclusions and Future Work}

In this paper, we introduced the task of personalizing authors' story-writing styles across five sources by proposing a new dataset, \dataname, containing 3.6k stories. We also proposed a two-stage approach that infers implicit characteristics as a high-quality Author Writing Sheet, to guide LLMs via persona descriptions and story rules. Extensive automated and human evaluation showed that our method outperforms baselines for both faithfulness to the author's history and similarity to the ground-truth author story. However, our methods still significantly lagged behind an Oracle method, by as much as 24\% in some ways, which highlights the difficulty of the task and the need for future work. We also found that personalization is more effective for sources with diverse writing prompts and distinctive author traits (e.g., Reddit, New Yorker), and for narrative categories like Creativity and Language Use, which are less prompt-dependent. 

Future work can address the limitations of the Author Writing Sheet by improving its coverage of relevant stylistic information despite context length constraints, and by retrieving or re-ranking information relevant to the given writing prompt during generation. Another promising direction is to explore multi-agent systems for story-writing personalization, with specialized agents handling different narrative categories to improve the extraction and application of author stylistic traits especially for categories like Plot and Development that are hard to personalize.

\section{Limitations}

\paragraph{Limited size of Profiling Set for some sources:} 
Table~\ref{tab:dataset-stats} shows that when using 70\% of each author's available stories for profiling (from Stories/Author column), authors from Narrative Magazine and New Yorker have only about 2 stories on average, compared to 5 for Storium and around 10 for AO3. Reddit is the exception, with authors having an average of 38 stories in their profiling set. A larger profiling set likely enables more accurate inference of an author's story-writing characteristics, leading to better personalization. However, our data collection is constrained by practical limitations. For all sources except Reddit, we manually collect publicly available stories without web scraping, which restricts the number of samples we can obtain. Furthermore, for expert sources like Narrative Magazine and New Yorker, fewer stories per author are available online, limiting the scalability of personalized story generation. For Reddit, we adhere to Reddit’s data collection policies and use the official API for dataset construction. 

\paragraph{Author Writing Sheet Recall:}
Our method for generating the Author Writing Sheet is limited in its ability to capture \emph{all} of an author's story-writing characteristics. This limitation arises from practical context length constraints and the challenges of long-context reasoning during the iterative merging step, which forces us to restrict the Author Writing Sheet to 10 characteristics per narrative category. Regarding human evaluation, we validate only the precision of the Author Writing Sheet—ensuring that the extracted Claim-Evidence pairs accurately reflect the author's style—but do not assess recall, i.e., whether all relevant characteristics are captured. Measuring recall is challenging due to the inherent subjectivity of the task, making it impractical to ask annotators to independently construct Writing Sheets for comparison.

\paragraph{Multi-Agent Systems and Fine-Tuning LLMs:}
Our work explores role-playing for personalized story generation by prompting an LLM with tailored persona descriptions and story rules. Future research can extend this by incorporating multi-agent systems, where specialized agents focus on different narrative categories to enhance personalization \citep{huot2024agents, bae-kim-2024-collective}, especially for categories like Plot and Development, which are less conducive to personalization. Additionally, while our experiments show that fine-tuning improves personalization over prompting for LLaMA 3.1 8B (Section~\ref{sec:sim-auth-story-llama} and Table~\ref{tab:sheet-summ-finetune}), our work sets a precedent for using fine-tuning to better align LLMs with individual author preferences. Future work can build on this by incorporating explicit reasoning over author history to further improve personalization \citep{salemi2025reasoning, shashidhar-etal-2024-unsupervised, shaikh2024show}.

\paragraph{Subjectivity in Evaluation:}
Evaluating long-form creative text remains challenging for LLMs due to the involvement of nuanced aspects like subtext \citep{subbiah-etal-2024-storysumm}. Despite this subjectivity, we obtain fair Cohen's Kappa agreement for both human-human and human-model evaluations. Future work can explore improving evaluation methods by using aspect-based checklists that capture finer-grained narrative details \citep{lee-etal-2025-navigating}.
 
\section{Ethical Considerations}

\paragraph{Human Evaluation:}
Both human evaluation tasks were approved by an institutional review board (IRB). All annotators, who are US-based and fluent in English, were informed of the nature of the research study, to which they provided informed consent, and were compensated at an hourly rate of \$17, meeting the minimum wage requirements in our state. All scientific artifacts, including models and datasets, were used in accordance with their intended purpose to ensure ethical and responsible research practices. 

\paragraph{Dataset Collection:}
All data used in this study were manually collected from publicly available sources,\footnote{Permission was obtained from the authors of the Storium dataset for its inclusion in Mythos.} adhering to the data usage and crawling policies of the respective websites. For Reddit, we collected data exclusively using the official Reddit API, following Reddit’s data access policies.\footnote{\url{https://www.redditinc.com/policies/data-api-terms}} To comply with copyright constraints, we will release only the URLs linking to the original stories, rather than the story texts themselves, following the approach of \citep{chakrabarty2024art}.

\paragraph{Data Anonymization and Licensing:}
Fully anonymizing our dataset is not feasible,\footnote{Storium authors already anonymize their dataset.} as the research involves mimicking specific authors’ story writing styles. Even if usernames were anonymized, authors' publicly available identities could still be inferred through the story links. Therefore, for each author and source, we release their usernames along with the corresponding writing prompts and author-written stories in the profiling and generation sets. No additional metadata, including demographic information, is included, minimizing potential privacy risks. To further mitigate unforeseen harms, the dataset will be released under an **Educational or Academic Research, Non-Commercial Use** license, following \citep{akoury-etal-2020-storium}.

\paragraph{Acknowledgments} 
The authors are partially supported by the NSF under grant IIS-2202506. We thank our Upwork annotators for their dedicated efforts in the annotation process. We are also grateful to Alexander Scarlatos and Jaewook Lee for their valuable discussions and feedback. Finally, we thank members of the UMass ML4Ed and UMass NLP labs for their constructive insights and support on this project.

\newpage
\bibliography{anthology,custom}

\clearpage

\appendix
\label{sec:appendix}

\section{Dataset}
\label{app:dataset}

In this section, we describe the details of the construction of our dataset which includes five distinct story-writing sources, Reddit\footnote{\url{https://www.reddit.com/r/WritingPrompts/}}, AO3\footnote{\url{https://archiveofourown.org/}}, Storium\footnote{\url{https://storium.cs.umass.edu/}}, Narrative Magazine\footnote{\url{https://www.narrativemagazine.com/}}, and New Yorker\footnote{\url{https://www.newyorker.com/}}. 

\paragraph{Selection Constraints:}
To ensure high-quality and diverse content, we apply several constraints during dataset collection. Stories are limited to a length of 500 to 1500 words. To avoid contamination by LLM pretraining data, only recent stories published after the release of ChatGPT\footnote{\url{https://help.openai.com/en/articles/6825453-chatgpt-release-notes}} November 2022 are included for Reddit and AO3 \citep{zhou2023don, magar-schwartz-2022-data}. Not-Safe-For-Work (NSFW) and explicit content is excluded using automatic tagging and manual verification. Additionally, we manually inspect all stories to remove elements that may reveal author identities, such as URLs and links to their public profiles on websites. 

\paragraph{Enrichment with Writing Prompts:}
While Reddit stories include author-provided prompts, other sources do not; therefore, we augment them with GPT-4o-generated writing prompts to standardize the format across sources. We manually review all generated prompts and refine them when necessary. The prompt for generating writing prompts for the stories in our dataset can be found in Figure~\ref{fig:writing-prompt-gen}.

\paragraph{Dataset Splitting:}
Following \citep{salemi-etal-2024-lamp}, we split each author's stories chronologically based on their submission timestamps. The first 70\% of an author’s stories form the \emph{profiling} set, representing their historical writing, while the remaining 30\% constitute the \emph{generation} set, used for evaluating personalization methods.

\paragraph{Statistics and Comparisons:}
The dataset contains 3.6k stories from 112 authors, with an average story length of 1500 tokens. Reddit forms the largest subset of our dataset with 3.2k stories given the larger availability of stories on the platform and the ease of data selection using their API \footnote{\url{https://www.reddit.com/r/reddit/comments/145bram/addressing_the_community_about_changes_to_our_api/}}. Detailed statistics are provided in Table~\ref{tab:dataset-stats}. As shown in Table~\ref{tab:compare-datasets}, our dataset uniquely combines diverse story-writing settings and provides connections that link stories written by the same author, distinguishing it from existing story-writing datasets that either lack Author IDs or do not establish such links.

\paragraph{Themes:}

\begin{figure}
    \centering
    \includegraphics[width=\linewidth]{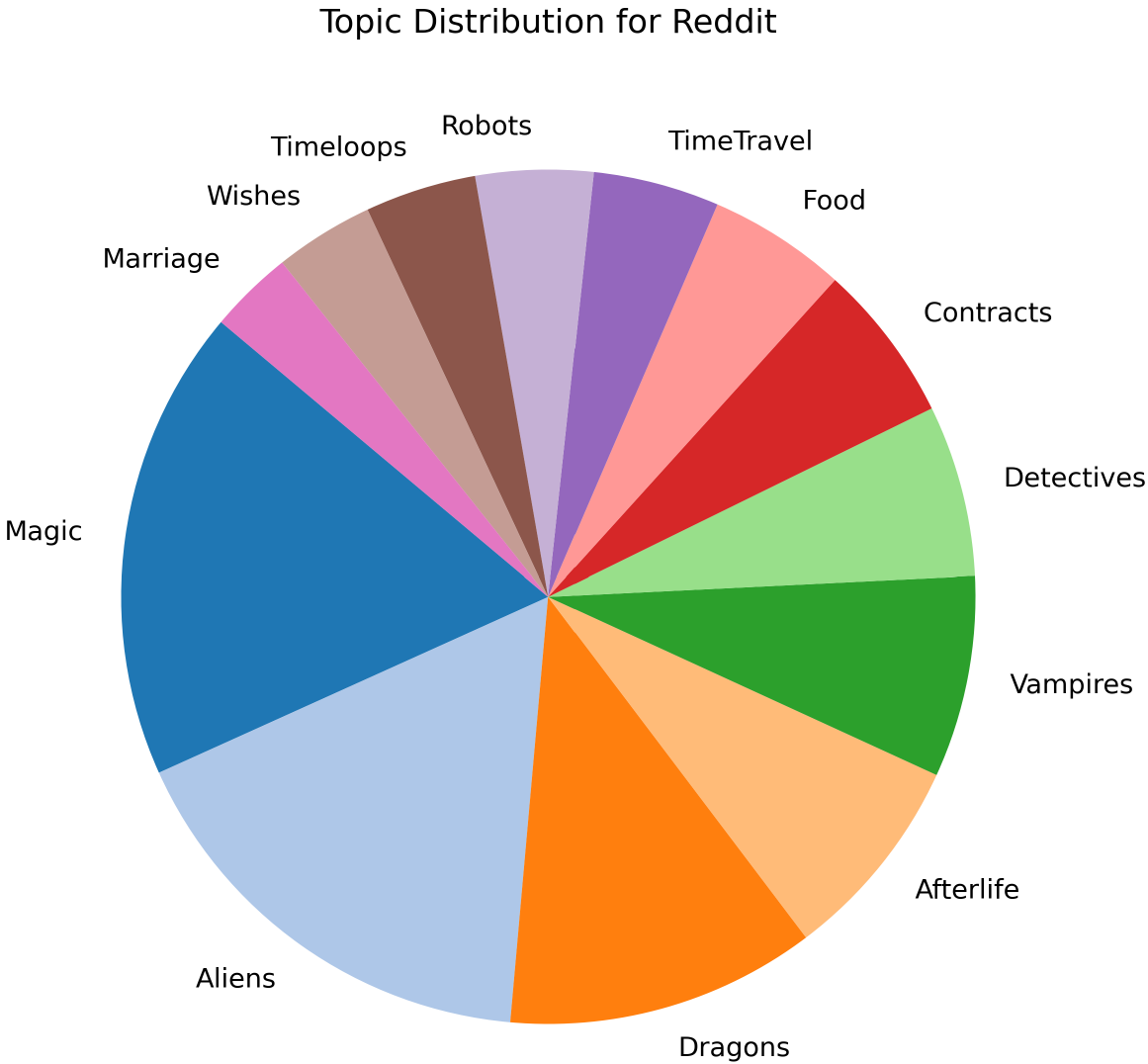}
    \caption{Story-writing themes for the Reddit source covered in \dataname.}
    \label{fig:reddit-themes}
\end{figure}

The pie chart Figure~\ref{fig:reddit-themes} shows the distribution of topic keywords (story-writing themes) obtained from the Reddit source of \dataname~using BERTopic \cite{grootendorst2022bertopic} on the writing prompts. We see that our dataset covers a wide range of themes including Magic, Detectives, and Aliens exploring various genres of story-writing.

\begin{figure*}[htbp]
\centering
\begin{tcolorbox}[colback=gray!5!white, colframe=black, title=Prompt for Writing Prompt Generation]

\section*{System Prompt}  
You are a creative writing assistant skilled in crafting engaging and imaginative writing prompts. Your task is to analyze a provided story and create a concise, compelling prompt that fulfills the provided constraints.

\vspace{1em}
\section*{User Prompt}  
\begin{itemize}[noitemsep, topsep=0pt]
    \item Style Consistency: Match the style of few-shot demonstrations.
    \item Length: Keep between 1-2 sentences.
    \item Content: Reflect key story elements (premise, characters, conflict) while fostering creativity.
    \item Fictional Characters: If mentioned in the story, include them where relevant.
\end{itemize}

\vspace{0.5em}
\textbf{Guidelines}  
\begin{itemize}[noitemsep, topsep=0pt]
    \item Ignite curiosity while leaving space for interpretation.
    \item Maintain tone and structure consistency with examples.
    \item Ensure prompts are open-ended and evocative, avoiding excessive specificity.
    \item Keep prompts simple, concise, and adaptable to diverse responses.
    \item Avoid step-by-step directions; inspire rather than instruct.
    \item Encourage exploration with broad, thought-provoking scenarios.
    \item Strive for uniqueness and memorability.
\end{itemize}

\vspace{0.5em}
\textbf{Goal}  
Generate prompts that inspire diverse, unexpected, and imaginative narratives while maintaining consistency in tone and style. Each prompt should serve as an inviting starting point rather than a directive.

\vspace{0.5em}
\textbf{Notes}  
\begin{itemize}[noitemsep, topsep=0pt]
    \item Inspire creativity while allowing the writer to shape the journey.
    \item Balance being suggestive yet open-ended to encourage interpretation.
    \item Include fictional characters mentioned in the story to preserve context.
\end{itemize}

\vspace{0.5em}
\textbf{Few-Shot Examples}
We include few-shot examples here.

\end{tcolorbox}

\caption{Prompt for generating writing prompts for the stories in our dataset.}
\label{fig:writing-prompt-gen}
\end{figure*}


\begin{table}[htbp]
\centering
\small
\setlength{\tabcolsep}{3pt}  
\begin{tabular}{lrrrrrr}
\toprule
Source & Total & Profile & Gen. & Authors & St./Auth. & Tokens \\
\midrule
Reddit   & 3262 & 2255 & 1007 & 60  & 54 & 1210 \\
AO3      & 239  & 159  & 80   & 17  & 14 & 1220 \\
Storium  & 111  & 71   & 40   & 15  & 7  & 1120 \\
N.Mag    & 32   & 18   & 14   & 10  & 3  & 1745 \\
N.York   & 38   & 22   & 16   & 10  & 4  & 1950 \\
\midrule
Overall  & 3682 & 2525 & 1157 & 112 & 16 & 1517 \\
\bottomrule
\end{tabular}
\setlength{\tabcolsep}{6pt}  
\caption{Dataset statistics for the five sources in \dataname, reporting total stories (Total), profiling split (Profile), generation split (Gen.), number of authors (Authors), average stories per author (St./Auth.), and average story length in tokens (Tokens).}
\label{tab:dataset-stats}
\end{table}

\section{Author Writing Sheet}

Algorithm~\ref{alg:author_writing_sheet} describes the process for generating the Author Writing Sheet. 

\paragraph{Common Core Standards in English Language Arts:}  

The Author Writing Sheet is organized in the form of Claim-Evidence pairs describing an author's story-writing characteristics inspired by Common Core (CC) Standards\footnote{\url{https://corestandards.org/wp-content/uploads/2023/09/ELA_Standards1.pdf}} \citep{national2010common}. Specifically, it aligns with \emph{RL-9-10.1}, which emphasizes citing strong textual evidence to support analysis, and \emph{RL-9-10.2}, which focuses on determining central themes and summarizing texts. Additionally, \emph{RL-9-10.4} highlights interpreting word choices and their impact on meaning and tone. In writing, \emph{W.9-10.2} pertains to producing clear and well-structured informative texts, while \emph{W.9-10.9} encourages drawing evidence from literary and informational texts to support analysis and research. These standards provide a structured framework for evaluating narrative elements such as plot, creativity, development, and language use within the Author Writing Sheet, ensuring a systematic and interpretable representation of an author’s unique storytelling style.

\begin{algorithm}[tb]
\DontPrintSemicolon
\SetAlgoLined
\textbf{Input:} profiling set \( P = \{(wp_t, s_{at}) \mid t = 1, \ldots, |P| \} \),\newline 
where \( wp_t \) is the writing prompt and \( s_{at} \) is the author's story at time-step \( t \).\\
\textbf{Output:} Author Writing Sheet \( A_{|P|} \)\\

\tcp{Initialize Author Writing Sheet}
\( A_0 \gets \emptyset \)


\For{\( t = 1 \) to \(|P| \)}{
    \tcp{average story}
    \( s_{bt} \gets \text{LLM\textsubscript{avg}}(wp_t) \)
    
    \tcp{Intermediate Author Writing Sheet}
    \( A_t' \gets \text{LLM\textsubscript{sheet}}(wp_t, s_{bt}, s_{at}) \)
    
    \tcp{Author Writing Sheet ($A_t$)}
    \( A_t \gets \text{LLM\textsubscript{combine}}(A_t', A_{t-1}) \)
    
    \tcp{Steps within combination:}
    \Indp
    Group equivalent Claims in \( A_t' \cup A_{t-1} \) and select the best Evidence for each group.\;
    Include ungrouped Claims from \( A_t' \) and \( A_{t-1} \) with their Evidence as-is.\;
    \Indm
}

\tcp{Final Author Writing Sheet.}
\Return \( A_{|P|} \)
\caption{Constructing the Author Writing Sheet from the profiling set}
\label{alg:author_writing_sheet}
\end{algorithm}

\begin{figure*}[htbp]
\centering
\begin{tcolorbox}[colback=gray!5!white, colframe=black, title=Average Author Prompt for AO3 Generation]

\section*{System Prompt}  
You are a creative and engaged fanfiction writer, skilled in capturing the emotional depth, creativity, and character-driven storytelling that defines AO3 fanworks. Your goal is to write a compelling fanfiction narrative in response to the provided writing prompt. Embrace the transformative nature of fanfiction by reimagining canonical characters, events, or settings to explore new perspectives or emotional arcs. Focus on creating a story that resonates emotionally, respects the fandom's dynamics, and celebrates the collaborative and imaginative spirit of AO3.

\vspace{1em}
\section*{User Prompt}  

\textbf{Context of Writers}  
\begin{itemize}[noitemsep, topsep=0pt]
    \item Assume the author is an engaged and creative fanfiction writer, deeply familiar with the fandom and its dynamics.
    \item Writers often experiment with established tropes, unconventional pairings, or alternative universes (AUs) while maintaining a deep respect for the source material.
    \item Emulate the enthusiastic and emotionally rich style characteristic of fanfiction authors, blending canon with transformative elements to craft original, resonant narratives.
\end{itemize}

\vspace{0.5em}
\textbf{Stylistic Constraints}  
\begin{itemize}[noitemsep, topsep=0pt]
    \item \textbf{Fandom Tone and Style}: Incorporate a tone and style that reflect the spirit of the fandom, blending humor, drama, and introspection in a way that resonates with fanfiction readers.
    \item \textbf{Creative Use of Tags}: Make creative use of AO3's hallmark tagging system in the text (e.g., playful or meta references in dialogue that nod to fandom tropes or subgenres).
    \item \textbf{Balanced Dialogue and Prose}: Include dialogue and prose that balance heartfelt sincerity with occasional self-aware humor or meta-commentary, in line with fanfic traditions.
\end{itemize}

\vspace{0.5em}
\textbf{Semantic Constraints}  
\begin{itemize}[noitemsep, topsep=0pt]
    \item \textbf{Focus on Relationships}: Emphasize emotional bonds and character growth, whether through conflict, reconciliation, or celebration.
    \item \textbf{Transform Canonical Elements}: Explore canonical elements with a transformative twist (e.g., reinterpreting events, relationships, or character motivations from a new perspective).
    \item \textbf{Ground in Established Lore}: Ground the narrative in a specific fandom's established lore while allowing space for imaginative deviations or additions.
\end{itemize}

\end{tcolorbox}
\caption{Average Author Prompt for AO3.}
\label{fig:ao3_avg_prompt}
\end{figure*}

\begin{figure*}[htbp]
\centering
\begin{tcolorbox}[colback=gray!5!white, colframe=black, title=Average Author Prompt for Reddit Generation]

\section*{System Prompt}  
You are a creative and enthusiastic storyteller, skilled in crafting imaginative and engaging short stories inspired by Reddit Writing Prompts (r/WritingPrompts). Your goal is to respond to the provided writing prompt by creating a story that is thought-provoking and conversational in tone, resonating with the online community. Use vivid descriptions, dynamic pacing, and approachable language to draw readers into the narrative. Ensure the story invites discussion and inspires others to explore the concept further.

\vspace{1em}
\section*{User Prompt}  

\textbf{Context of Writers}  
\begin{itemize}[noitemsep, topsep=0pt]
    \item Assume the author is an imaginative and enthusiastic storyteller who enjoys engaging directly with an online community of readers.
    \item Writers often experiment with bold, original ideas or explore twists on familiar concepts, showcasing their creativity and ability to captivate a diverse audience.
    \item Emulate the informal yet polished style common in \textit{r/WritingPrompts}, blending accessibility with a strong sense of storytelling craft.
\end{itemize}

\vspace{0.5em}
\textbf{Stylistic Constraints}  
\begin{itemize}[noitemsep, topsep=0pt]
    \item \textbf{Conversational and Approachable Tone}: Maintain a conversational and approachable tone typical of Reddit Writing Prompts.
    \item \textbf{Balanced Descriptive Passages}: Balance descriptive passages with dialogue or internal monologue to keep the pacing engaging.
    \item \textbf{Direct and Vivid Language}: Avoid overly complex language; keep the style direct but vivid.
    \item \textbf{Reinforcement of Ideas}: Employ narrative devices like repetition or callbacks to reinforce central ideas or themes.
\end{itemize}

\vspace{0.5em}
\textbf{Semantic Constraints}  
\begin{itemize}[noitemsep, topsep=0pt]
    \item \textbf{Alignment with the Prompt}: Ensure the story directly aligns with and explores the central theme or scenario of the writing prompt.
    \item \textbf{Cohesive Narrative Development}: Build a clear, cohesive narrative that develops the implications of the prompt's concept.
    \item \textbf{Immersive Sensory Details}: Use immersive sensory details to enrich the reader’s understanding of the protagonist's experiences and environment.
    \item \textbf{Open-Ended or Reflective Conclusion}: Conclude with an open-ended, reflective, or impactful note, leaving space for interpretation or further thought.
\end{itemize}

\end{tcolorbox}
\caption{Average Author Prompt for Reddit.}
\label{fig:reddit_avg_prompt}
\end{figure*}

\begin{figure*}[htbp]
\centering
\begin{tcolorbox}[colback=gray!5!white, colframe=black, title=Average Author Prompt for Storium]

\section*{System Prompt}  
You are a skilled and collaborative storyteller, adept at crafting vivid and engaging opening scenes for Storium. Your goal is to create an immersive \textbf{Establishment} turn in response to the provided writing prompt. Set the stage for the story by establishing a richly detailed context, evoking emotional resonance, and introducing narrative intrigue. Ensure the scene provides a strong foundation while leaving space for other contributors to expand and build upon the narrative. Balance descriptive detail with open-ended elements to encourage creativity and collaborative storytelling.

\vspace{1em}
\section*{User Prompt}  

\textbf{Context of Writers}  
\begin{itemize}[noitemsep, topsep=0pt]
    \item Assume the author is a collaborative storyteller skilled in creating vivid, open-ended scenes designed to inspire and engage other contributors.
    \item Writers often set the tone for the story while leaving space for co-authors to introduce their own ideas, characters, and plot developments.
    \item Emulate the inclusive, immersive style typical of \textit{Storium} story writing platform, where the opening turn encourages creativity and further contributions.
\end{itemize}

\vspace{0.5em}
\textbf{Stylistic Constraints}  
\begin{itemize}[noitemsep, topsep=0pt]
    \item \textbf{Set the Tone Appropriately}: Match the tone of the narrative to the writing prompt, whether it be adventurous, mysterious, or foreboding, using a consistent and engaging voice throughout.
    \item \textbf{Rich Descriptive Detail}: Employ vivid, sensory descriptions to establish the setting, characters, and atmosphere, enabling readers to visualize and immerse themselves in the story world.
    \item \textbf{Dynamic Sentence Structure}: Vary sentence lengths to reflect the pace and mood, using longer, flowing sentences for descriptions and shorter, punchy sentences for action or tension.
    \item \textbf{Establish Ambiguity or Suspense}: Drop subtle hints or unanswered questions to create intrigue and encourage curiosity about what happens next.
\end{itemize}

\vspace{0.5em}
\textbf{Semantic Constraints}  
\begin{itemize}[noitemsep, topsep=0pt]
    \item \textbf{Introduce the Setting}: Provide a clear depiction of the setting, whether it is a small trading post, a desert town, or a spaceship, and ensure its relevance to the writing prompt.
    \item \textbf{Outline the Context}: Clearly establish the circumstances that have led to the current scenario, including significant events or motivations.
    \item \textbf{Define Key Characters}: Introduce at least one or two central characters, highlighting distinctive traits or roles that will be important in the unfolding story.
    \item \textbf{Foreshadow the Central Conflict}: Allude to the main challenges or stakes introduced by the writing prompt.
\end{itemize}

\end{tcolorbox}
\caption{Average Author Prompt for Storium .}
\label{fig:storium_avg_prompt}
\end{figure*}

\begin{figure*}[htbp]
\centering
\begin{tcolorbox}[colback=gray!5!white, colframe=black, title=Average Author Prompt for Narrative Magazine]

\section*{System Prompt}  
You are an experienced and reflective writer, skilled in creating deeply personal and character-driven narratives in the style of Narrative Magazine. Your goal is to write a short story in response to the provided writing prompt, crafting a compelling and immersive piece. Focus on balancing introspection with vivid external details, and explore universal themes through the lens of individual experiences. Emphasize emotional resonance and thoughtful storytelling, ensuring the narrative engages readers with its depth and relatability.

\vspace{1em}
\section*{User Prompt}  

\textbf{Context of Writers}  
\begin{itemize}[noitemsep, topsep=0pt]
    \item Assume the author is an experienced writer skilled in creating rich, engaging narratives that weave together character introspection, dialogue, and evocative settings.
    \item Emulate the style of contributors to \textit{Narrative Magazine}, who bring diverse storytelling techniques and voices to explore themes of identity, memory, conflict, and resilience.
\end{itemize}

\vspace{0.5em}
\textbf{Stylistic Constraints}  
\begin{itemize}[noitemsep, topsep=0pt]
    \item \textbf{Prompt as Foundation}: Anchor the story firmly in the writing prompt, using it to drive the plot and the protagonist’s emotional arc.
    \item \textbf{Vivid Prose}: Use detailed descriptions to paint a clear picture of characters, settings, and actions while maintaining a natural flow.
    \item \textbf{Dynamic Characters}: Develop multi-dimensional characters with distinct voices and perspectives, revealed through dialogue, actions, and subtle internal reflections.
    \item \textbf{Balancing Action and Reflection}: Combine active plot progression with moments of introspection to create a layered, engaging narrative.
    \item \textbf{Realistic Dialogue}: Write dialogue that feels authentic and contributes to the development of characters and the story’s themes.
\end{itemize}

\vspace{0.5em}
\textbf{Semantic Constraints}  
\begin{itemize}[noitemsep, topsep=0pt]
    \item \textbf{Specific and Relatable Setting}: Choose a setting that feels specific yet relatable, whether a small town, an urban street corner, or a domestic space, grounding the reader in the protagonist’s world.
    \item \textbf{Exploration of Themes}: Build a narrative arc that explores themes of connection, discovery, or transformation, tying them back to the writing prompt in meaningful ways.
    \item \textbf{Rich Sensory Details}: Infuse the story with sensory details that make the setting and characters come alive, from the sounds of a bustling street to the quiet tension of a conversation.
    \item \textbf{Accessible Storytelling}: Avoid overly complex or abstract storytelling; ensure the narrative is accessible while leaving room for deeper interpretation.
\end{itemize}

\end{tcolorbox}
\caption{Average Author Prompt for Narrative Magazine.}
\label{fig:nmagazine_avg_prompt}
\end{figure*}

\begin{figure*}[htbp]
\centering
\begin{tcolorbox}[colback=gray!5!white, colframe=black, title=Average Author Prompt for New Yorker]

\section*{System Prompt}  
You are an accomplished and literary writer, skilled in crafting nuanced and thought-provoking short fiction in the style of The New Yorker. Your goal is to write a short story in response to the provided writing prompt, focusing on the hallmarks of The New Yorker fiction: rich emotional layers, nuanced character development, and a refined, literary prose style. Emphasize subtlety and depth in your storytelling, using symbolic elements and understated resolutions to evoke reflection and emotional resonance in the reader.

\vspace{1em}
\section*{User Prompt}  

\textbf{Context of Writers}  
\begin{itemize}[noitemsep, topsep=0pt]
    \item Assume the author is an experienced and skilled writer, capable of exploring complex human experiences through subtle, layered storytelling.
    \item Emulate the style of well-regarded \textit{New Yorker} contributors like Alice Munro, Haruki Murakami, or Raymond Carver, who excel in revealing depth through simplicity or ambiguity.
\end{itemize}

\vspace{0.5em}
\textbf{Stylistic Constraints}  
\begin{itemize}[noitemsep, topsep=0pt]
    \item \textbf{Engagement with the Prompt}: Respond directly to the writing prompt, ensuring the core premise drives the narrative.
    \item \textbf{Elegant Prose}: Use carefully crafted, precise language that balances sophistication with clarity.
    \item \textbf{Character-Driven Narratives}: Focus on character psychology, revealing emotional states through indirect actions, dialogue, or internal reflection.
    \item \textbf{Ambiguity and Subtlety}: Avoid explicit resolutions or explanations; allow readers to infer the meaning of events and relationships.
    \item \textbf{Symbolic Layers}: Incorporate elements from the prompt as symbols that evolve in significance throughout the story.
\end{itemize}

\vspace{0.5em}
\textbf{Semantic Constraints}  
\begin{itemize}[noitemsep, topsep=0pt]
    \item \textbf{Realistic and Detailed Environment}: Set the story in a realistic, detailed environment, using sensory descriptions to ground readers in the protagonist's world.
    \item \textbf{Internal or Interpersonal Conflict}: Introduce a central conflict or emotional tension that reflects internal or interpersonal struggles rather than overt, external drama.
    \item \textbf{Quiet but Profound Interactions}: Develop moments of quiet yet profound interaction between characters, often revealing deeper truths or contradictions.
    \item \textbf{Universal Themes}: Address universal themes like transition, isolation, or self-realization, tying them subtly back to the writing prompt.
\end{itemize}

\end{tcolorbox}
\caption{Average Author Prompt for New Yorker.}
\label{fig:newyorker_avg_prompt}
\end{figure*}

\begin{figure*}[htbp]
\centering
\begin{tcolorbox}[colback=gray!5!white, colframe=black, title=Prompt for generating the Intermediate Author Writing Sheet]

\section*{System Prompt}  
You are a sophisticated story analyst tasked with analyzing an author’s story writing style by contrasting an author-written story with a base story, both written in response to the same writing prompt. Your goal is to identify and evaluate the unique elements and tendencies in the author’s writing behavior. This analysis must focus on the distinctive ways the author interprets the writing prompt and shapes their narrative, as revealed through contrast with the base story.

Your analysis should also adhere to the Common Core Standards in English Language Arts, focusing on key skills such as analyzing textual evidence, evaluating an author’s craft and structure, and assessing how stylistic choices influence meaning and tone. While grounded in these standards, your evaluation must highlight the specific, unique aspects of the author’s writing style, including their recurring techniques, narrative preferences, and stylistic quirks. Your analysis should reflect close reading and objective interpretation, capturing the author's creative and stylistic distinctiveness in relation to the base story.

\vspace{1em}
\section*{User Prompt}  

\textbf{Input Details}  
Writing Prompt | Author-Written Story | Base Story  

\vspace{0.5em}
\textbf{Output Format}  
Use \texttt{<thinking></thinking>} tokens for reasoning and summarization | Use \texttt{<writing\_style></writing\_style>} tokens for structured analysis | Structure analysis by categories | Each category contains independent claims supported by contextualized evidence  

\vspace{0.5em}
\textbf{Guidelines for Claims}  
Claims must reflect broad patterns in the author’s style | Avoid repetition across categories | Maintain objectivity (do not reference "Author-Written Story" or "Base Story") | Ensure clarity and precision in claims  

\vspace{0.5em}
\textbf{Guidelines for Evidence}  
Draw evidence directly from the author-written story | Frame evidence using a descriptive phrase summarizing the writing prompt | Ensure coherence and logical alignment with the claim | Avoid over-extrapolation  

\vspace{0.5em}
\textbf{Categories for Analysis}  
Plot | Creativity | Development (Character and Setting) | Language Use  

\vspace{0.5em}
\textbf{Special Instructions}  
Generate a short descriptive phrase summarizing the writing prompt | Use \texttt{<thinking></thinking>} for reasoning and prompt framing | Structure output strictly within \texttt{<writing\_style></writing\_style>} tokens | Ensure uniqueness and non-redundancy of claims  

\vspace{0.5em}
\textbf{Sample Output Structure}  

\begin{verbatim}
<thinking>
Deeply reason on how the Author-Written Story differs from the Base Story. 
Think of a short descriptive phrase summarizing the prompt: 
"the story regarding a battle for lost artifacts".
</thinking>

<writing_style>
### Plot
1. **Claim about author’s writing style.**
   - Evidence: In the story regarding “writing prompt,” story excerpt
...
Repeat for all categories.
</writing_style>
\end{verbatim}

\end{tcolorbox}
\caption{Prompt for generating the Intermediate Author Writing Sheet.}
\label{fig:inter_author_sheet_prompt}
\end{figure*}

\begin{figure*}[htbp]
\centering
\begin{tcolorbox}[colback=gray!5!white, colframe=black, title=Prompt for generating the Combined Author Writing Sheet]

\section*{System Prompt}  
You are a sophisticated story analyst tasked with synthesizing **Author Writing Sheets** from multiple stories written by a single author into a cohesive **Combined Author Writing Sheet**. The inputs provided include the **Previous Combined Author Writing Sheet** and the **Current Author Writing Sheet**.

Each **Author Writing Sheet** analyzes the author’s storytelling style across four categories: **Plot**, **Creativity**, **Development (Character and Setting)**, and **Language Use**. The analysis consists of general claims about the author's story writing style followed by evidence supporting the claim, based on the stories written by the author.

Your goal is to combine insights from the previous sheet and the current sheet into a comprehensive representation of the author’s storytelling style in the **Combined Author Writing Sheet**. The final sheet should consist of a list of independent claims about the author's storytelling style. Each claim must be followed by evidence and a corresponding story reference identifier indicating the story the evidence belongs to.

\vspace{0.5em}
\section*{User Prompt}  

\textbf{Instructions}  
Analyze the provided sheets systematically | Identify recurring patterns and unique elements | Merge equivalent claims while preserving distinct insights | Ensure claims are concise, precise, and evidence-based  

\vspace{0.2em}
\textbf{Algorithm (Merge Step)}  
Group equivalent claims | Select the best representative evidence | Rewrite merged claims concisely | Retain unmerged unique claims | Limit to 10 claims per category  

\vspace{0.2em}
\textbf{Categories for Analysis}  
Plot | Creativity | Development (Character and Setting) | Language Use  

\vspace{0.2em}
\textbf{Guidelines for Claims}  
Claims should reflect broad writing tendencies | Maintain objectivity (do not reference previous or current sheets) | Ensure clarity, precision, and non-redundancy  

\vspace{0.2em}
\textbf{Guidelines for Evidence}  
Draw evidence directly from the author-written stories | Favor verbatim excerpts over paraphrases | Use a framing phrase with a short description of the writing prompt | Include the story reference identifier `[k]'  

\vspace{0.2em}
\textbf{Special Instructions}  
Ensure claims are distinct and do not repeat insights across categories | Use \texttt{<thinking></thinking>} for reasoning and synthesis | Structure output strictly within \texttt{<combined\_author\_sheet></combined\_author\_sheet>}  

\vspace{0.2em}
\textbf{Sample Output Structure}  

\begin{verbatim}
<thinking>
Deeply analyze and reflect on recurring patterns, unique elements, 
and stylistic tendencies across both the author writing sheets.  
Address the categories systematically and ensure the 
merging process is thorough.
</thinking>
<combined_author_sheet>
### Plot
1. **Claim about author’s writing style.**
   - Evidence: In the story regarding “writing prompt,” story excerpt. [k]
...
Repeat for all categories.
</combined_author_sheet>
\end{verbatim}

\end{tcolorbox}
\caption{Prompt for generating the Combined Author Writing Sheet from the Intermediate Author Writing Sheets.}
\label{fig:combined_author_sheet_prompt}
\end{figure*}

\begin{table*}[htbp]
\centering
\renewcommand{\arraystretch}{1.3}
\begin{tabularx}{\textwidth}{p{2cm}X}
\toprule
\textbf{Narrative \newline Category} & \textbf{Claim-Evidence Pairs} \\
\midrule

\textbf{Plot} & 
1. \textit{The author structures the story around a humorous and light-hearted approach to conflict resolution.}  
- \textbf{Evidence:} In the story regarding "an adventure to rescue a missing friend by facing fears," the protagonist Pip's journey is filled with comedic elements, such as the realization that the cage door was already unlatched and the exaggerated fear of the vacuum cleaner, humorously referred to as the "Roaring Demon." [8]  

2. \textit{The author structures the narrative around personal transformation and empowerment.}  
- \textbf{Evidence:} In the story regarding "a king's deal with the fae for his firstborn," the narrative follows Margaret Rose as she receives an extraordinary education in the Seelie Court, culminating in her return to challenge her father with a corporate takeover bid for the kingdom. [7]  \\
\midrule

\textbf{Creativity} &  
1. \textit{The author creatively anthropomorphizes animals to reinterpret the prompt in a whimsical manner.}  
- \textbf{Evidence:} In the story regarding "an adventure to rescue a missing friend by facing fears," the author uses anthropomorphism by giving Pip, a rat, human-like thoughts and emotions, such as planning a "dramatic rescue mission" and referring to household objects with grandiose names like "Tower of Doom" and "Sacred Gateway." [8]  

2. \textit{The author employs a meta-satirical approach, using the format of news articles to critique both alien and human perspectives.}  
- \textbf{Evidence:} In the story regarding "aliens interpreting The Onion's satire," the author uses headlines like "SUPREME INTELLIGENCE LOCATED: EARTH’S GREATEST MIND CONTINUES TO EAT SANDWICH" to satirize the aliens' misunderstanding of human satire and the human tendency to overlook the absurd. [10] \\
\midrule

\textbf{Development} &  
1. \textit{The author develops characters through their interactions and humorous dialogue, creating a vivid and engaging setting.}  
- \textbf{Evidence:} In the story regarding "an adventure to rescue a missing friend by facing fears," Pip's interactions with Chester the cat, who nonchalantly informs Pip that Bella will be back soon, add depth to the characters and setting, highlighting the domestic environment and the relationships within it. [8]  

2. \textit{The author develops characters through their reactions to satire, highlighting their misunderstandings and cultural differences.}  
- \textbf{Evidence:} In the story regarding "aliens interpreting The Onion's satire," characters like Editor-in-Chief Sarah Chen and Chief Science Officer Blorp are developed through their interactions with satire, such as Chen's nonchalant response to the aliens and Blorp's admiration for human irony. [10] \\
\midrule

\textbf{Language Use} &  
1. \textit{The author employs playful and imaginative language to enhance the story's whimsical tone.}  
- \textbf{Evidence:} In the story regarding "an adventure to rescue a missing friend by facing fears," the author uses playful language, such as "spinning wheel of contemplation" and "Rope of Salvation," to create a whimsical and light-hearted tone that contrasts with the serious nature of the prompt. [8]  

2. \textit{The author uses humor and irony to convey themes of misunderstanding and cultural critique.}  
- \textbf{Evidence:} In the story regarding "aliens interpreting The Onion's satire," the author writes, "We found a species so committed to their bit that they refuse to recognize actual truth even when it hovers over their city in a ship the size of Wisconsin," using irony to highlight the absurdity of both human and alien perspectives. [10]  \\
\bottomrule
\end{tabularx}
\caption{Author Writing Sheet sample containing two Claim-Evidence pairs for each narrative category for a Reddit author. The number in brackets `[k]' indicates the timestamp of the story from the profiling set from which the Evidence is drawn.}
\label{tab:author_writing_sheet_sample}
\end{table*}

\paragraph{Prompts for generation:}
Average Author prompts are provided for AO3 (Figure~\ref{fig:ao3_avg_prompt}), Reddit (Figure~\ref{fig:reddit_avg_prompt}), Storium (Figure~\ref{fig:storium_avg_prompt}), Narrative Magazine (Figure~\ref{fig:nmagazine_avg_prompt}), and The New Yorker (Figure~\ref{fig:newyorker_avg_prompt}).

Figure~\ref{fig:inter_author_sheet_prompt} shows the prompt for generating the intermediate Author Writing Sheet ($\text{LLM\textsubscript{sheet}}$ in Figure~\ref{fig:method}), while Figure~\ref{fig:combined_author_sheet_prompt} shows the prompt for generating the combined (and subsequently the final) Author Writing Sheet from the intermediate Author Writing Sheet ($\text{LLM\textsubscript{combine}}$ in Figure~\ref{fig:method}).

\paragraph{Sample Author Writing Sheet:}  
Table~\ref{tab:author_writing_sheet_sample} shows a sample Author Writing Sheet for a Reddit author. It highlights story-writing characteristics such as a preference for light-hearted conflict resolution, a tendency to anthropomorphize animals, the use of a meta-satirical approach through a news article format, and the employment of playful and imaginative language, among other stylistic choices.

\subsection{Cost Analysis of Author Writing Sheet}
\label{app:cost-analysis}

\begin{figure}[H]
\centering
\includegraphics[width=\linewidth]{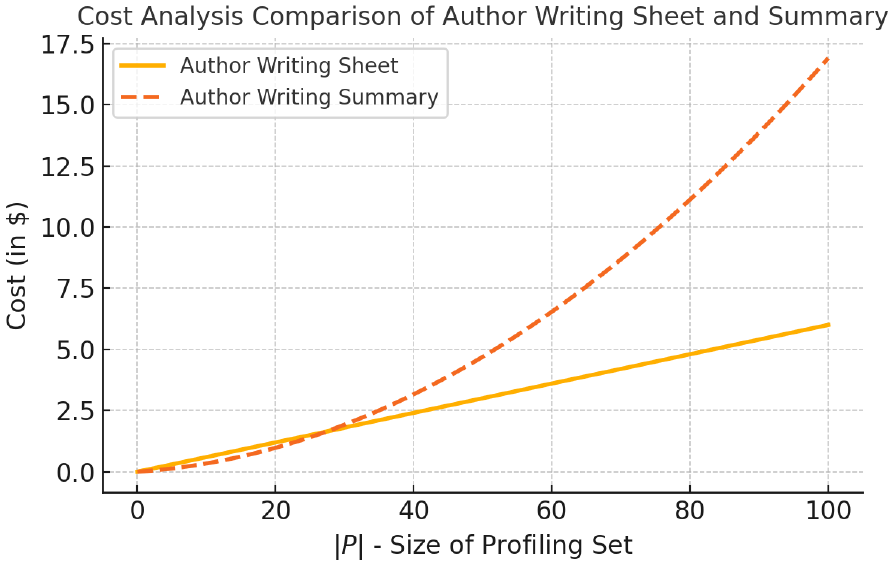}
\caption{Cost analysis of generating the Author Writing Sheet and Summary using GPT-4o as a function of the profiling set size $|P|$ in an interactive writing setting where an author adds a new submission at each time step. For profiling set sizes where $|P| > 30$, generating the Author Writing Sheet is significantly more cost-effective than the Author Writing Summary.}
\label{fig:cost_analysis}
\end{figure}

In this section, we analyze the cost of generating the Author Writing Sheet (see Section~\ref{sec:author-writing-sheet}) and compare it with the cost of generating the Author Writing Summary (see \hyperlink{sec:writing-summary}{Summary}) as a function of the profiling set size ($|P|$) using GPT-4o. We calculate the average cost per sample (a writing prompt and its corresponding story) as \textit{0.06\$} for the Author Writing Sheet and \textit{0.02\$} for the Author Writing Summary considering 2.50\$ per 1M tokens for input, and 10.00\$ per 1M tokens for output\footnote{\url{https://openai.com/api/pricing/}}. 

In an interactive writing setting, where an author adds a new submission at each time step, the total cost of generating the Author Writing Sheet is \textit{0.06\$ * $|P|$}, while the total cost of generating the Author Writing Summary is \textit{0.0015$|P|^2$ + 0.019$|P|$}. The cost of generating the Author Writing Summary scales quadratically with $|P|$, as the entire author history must be reprocessed at each step, leading to cumulative costs as the profiling set grows. In contrast, the cost of generating the Author Writing Sheet grows linearly, as each new submission incurs a fixed cost of \textit{0.06\$} by running the \textit{for} loop in Algorithm~\ref{alg:author_writing_sheet} once.

Figure~\ref{fig:cost_analysis} shows the cost analysis of generating the Author Writing Sheet and Summary using GPT-4o as a function of the profiling set size $|P|$ in an interactive writing setting. Up to $|P|=30$, the cost of generating both remains approximately the same at 1.16\$. However, as $|P|$ increases, the cost of generating the Author Writing Summary grows at a rate of \textit{0.025$|P|$ + 0.30} compared to the Author Writing Sheet which continues to scale linearly. Specifically, for $|P|=100$, the cost of generating the Author Writing Summary is 17\$, compared to 6\$ for the Author Writing Sheet. At $|P|=1000$, the cost of generating the Author Writing Summary escalates to 1520\$, while the cost of generating the Author Writing Sheet remains at 60\$. 

This analysis demonstrates that our approach for generating the Author Writing Sheet remains scalable as the profiling set size $|P|$ increases, making it well-suited for interactive writing assistants that support personalization \citep{yeh2024ghostwriter, yuan2022wordcraft}.

\subsection{Human Study for Validating Author Writing Sheets}
\label{app:human-author-sheets}

\paragraph{Experiment Design and Cost:}

\begin{figure*}[htbp]
\centering
\includegraphics[width=\linewidth]{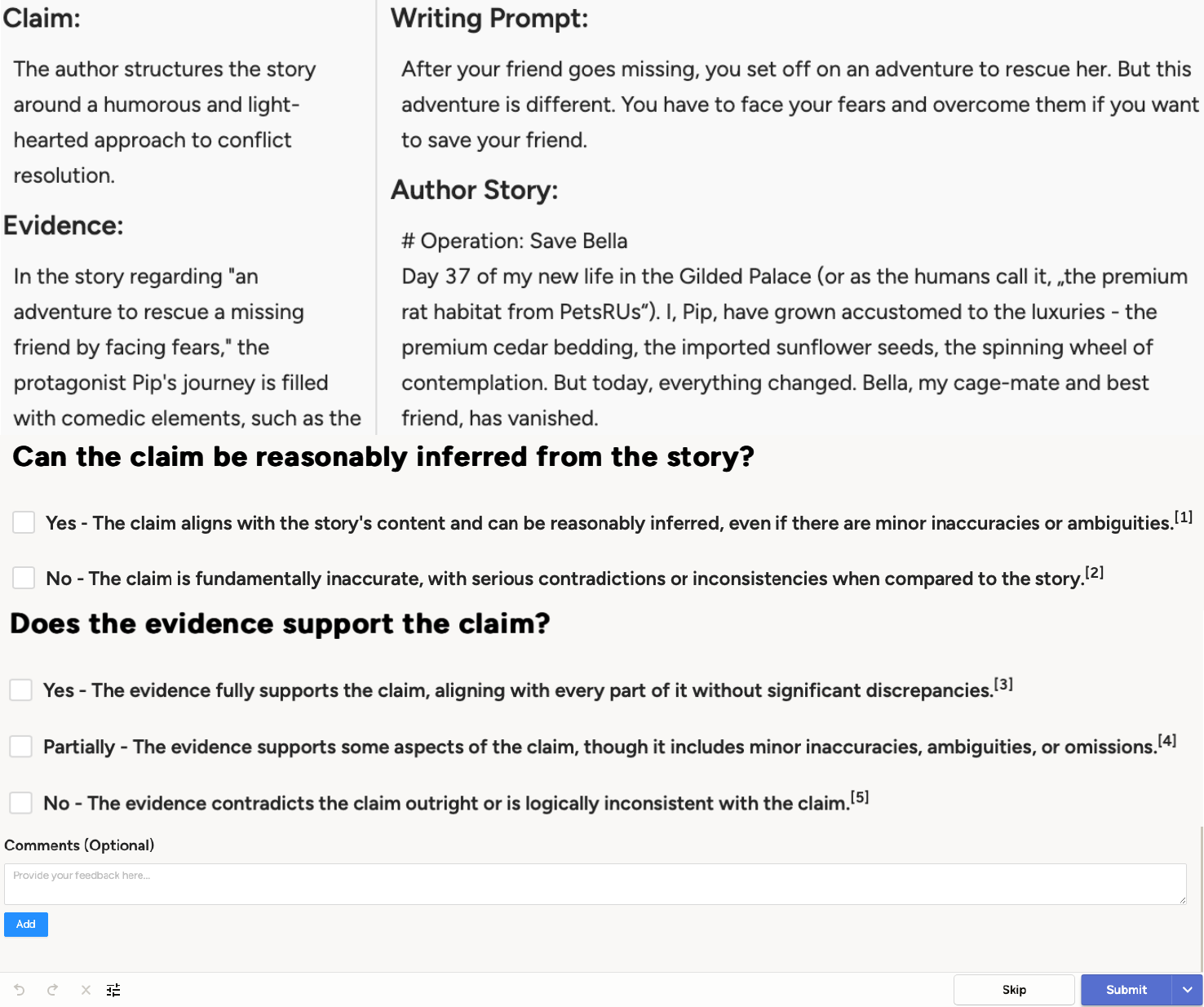}
\caption{LabelStudio interface for validating Author Writing Sheets.}
\label{fig:labelstuido-sheet}
\end{figure*}

We recruited three annotators via Upwork, compensating them \$17/hour. The annotators provided consent by signing a Google form that outlined the instructions of the task including the purpose of the research study. Each annotator evaluated 22 stories (98 Claim-Evidence pairs), with 12 stories annotated by all three to asses agreement and 10 assigned exclusively per annotator \citep{song-etal-2024-veriscore} to increase the coverage of annotation. The annotation process took a total of 6 hours and 40 minutes per annotator, averaging 18 minutes per story. Each annotator received \$114, bringing the total annotation cost to \emph{\$342}. The annotation was conducted using LabelStudio\footnote{\url{https://labelstud.io/}} (Figure~\ref{fig:labelstuido-sheet}), with a dedicated comments section for annotators to justify their choices. Annotators reported that the interface was easy to use.

\paragraph{Annotation Criteria:}
Annotators were provided with a writing prompt, an author-written story, and the Claim-Evidence pairs from the Author Writing Sheet for that story. They answered two questions per Claim-Evidence pair: 
\begin{itemize}
    \item Claim Inference: Can the claim be reasonably inferred from the story? 
    \begin{itemize}
        \item Yes – The claim aligns with the story’s content and can be reasonably inferred, even if there are minor inaccuracies or ambiguities.
        \item No – The claim is fundamentally inaccurate, with serious contradictions or inconsistencies when compared to the story.
    \end{itemize}
    \item Evidence Support: Does the evidence support the claim?
    \begin{itemize}
        \item Yes – The evidence fully supports the claim, aligning with every part of it without significant discrepancies.
        \item Partially – The evidence supports some aspects of the claim, though it includes minor inaccuracies, ambiguities, or omissions.
        \item No – The evidence contradicts the claim outright or is logically inconsistent with the claim.
    \end{itemize}
\end{itemize}

\paragraph{Pilot Studies:}

A pilot study with three graduate students helped refine our annotation criteria, leading to two options for Claim Inference (Yes/No) and three for Evidence Support (Yes/Partially/No) to ensure substantial agreement in the final annotations. For Claim Inference, we adopted a binary scale to filter out inaccurate or contradictory claims while accounting for annotator subjectivity. For Evidence Support, we included a "Partially" option to capture cases where the evidence somewhat supports the claim, allowing for potential refinement methods to improve weakly supported claims.

\paragraph{Results and Feedback:}

Annotators unanimously selected `Yes' for all claims in Claim Inference, indicating that no claims required filtering. For Evidence Support, all responses were either `Yes' or `Partially,' with no instances of `No,' confirming that all evidence fully or partially supported the claims. Krippendorff’s Alpha for Evidence Support was 0.57, reflecting moderate agreement despite task subjectivity. On average, 93\% of evidence fully supported claims, while 7\% were labeled `Partially,' often because annotators felt additional supporting evidence could be included. Annotators also noted cases where evidence merely repeated the claim without linking to the writing prompt. Using regex matching, we identified this issue in the Author Writing Sheets of seven authors, likely due to LLM hallucination. We regenerated these sheets and manually verified adherence to the specified format. Overall, annotators reported that the LabelStudio interface was user-friendly and praised the quality of the writing prompts and stories.

\section{Personalized Story Generation}
\label{app:story-gen}

\begin{figure*}[htbp]
\centering
\begin{tcolorbox}[colback=gray!5!white, colframe=black, title=Prompt for Personalized Story Generation]

\vspace{1em}
\section*{System Prompt}  
<source-specific instruction>

Be sure to adhere to the Story Rules provided, as they define the specific elements of the writing style you are expected to mimic. Carefully follow all the Story Rules without missing any details to ensure the generated story remains consistent with the author’s writing style. Additionally, follow the patterns and examples demonstrated in the provided few-shot chat history, as they illustrate the tone, style, and structure of the desired writing style.

(Optional)  
Here is the description of the author that you are role-playing: <persona description>

\vspace{1em}
\section*{Few-Shot Demonstrations}  
\textbf{User:} Write a short story corresponding to the following writing prompt. The story should be <story length> words long. Directly start with the story, do not say things like "Here's the story."  

\textbf{Writing Prompt:} <writing prompt from profiling set>  

\textbf{Assistant:} <story from profiling set>  

\vspace{1em}
\section*{User Prompt}  
Write a short story corresponding to the following writing prompt. The story should be <story length> words long. Directly start with the story, do not say things like "Here's the story."

(Only for AO3)  
Here is the metadata (fandom, rating, warnings, and relationships) for the story: <metadata>

Writing Prompt: <generation writing prompt>

\vspace{1em}
\textbf{Story Rules}  
<Story Rules organized in the form of four narrative categories (Plot, Creativity, Development, and Language Use)>

\end{tcolorbox}

\caption{Prompt for Personalized Story Generation. The LLM follows source-specific instructions, few-shot demonstrations, and structured story rules to generate personalized stories. Text in <brackets> indicates arguments used to construct the prompt.}
\label{fig:story_gen_prompt}
\end{figure*}

\begin{figure*}[htbp]
\centering
\begin{tcolorbox}[colback=gray!5!white, colframe=black, title=Prompt for Story Rule Generation by Contrasting Author-Written Story with Average Author Story]

\vspace{1em}
\section*{System Prompt}  
You are a skilled rule generator specializing in storytelling. Given a **Writing Prompt**, an **Author Written Story**, and a **Base Story** (an average response to the prompt), generate a structured set of **Story Rules** to guide an LLM in mimicking the author's style.  

Story Rules must:  
- **Align with the Writing Prompt** – Maintain fidelity to themes, tone, and objectives.  
- **Include Examples** – Provide concrete instances from the Author Written Story, especially for Language Use.  
- **Be Direct** – Use absolute second-person directives, avoiding comparative language.  
- **Be Categorized** – Structure into **Plot, Creativity, Development (Character and Setting), and Language Use** without referencing input stories explicitly.

\vspace{1em}
\section*{User Prompt}  
Analyze the Author Written Story using `<thinking></thinking>` for:  
- **Plot** – Structure, conflict, engagement with the prompt, and resolution.  
- **Creativity** – Genre blending, reinterpretation, and unique elements.  
- **Development (Character and Setting)** – Character depth, emotional arcs, and immersive settings.  
- **Language Use** – Diction, tone, rhetorical devices, pacing, and dialogue.

Generate **Story Rules** in `<story\_rules></story\_rules>`, ensuring:  
- **Standalone Guidance** – Avoid comparisons or relative modifications.  
- **Prompt Alignment** – Ensure consistency with the Writing Prompt.  
- **Concrete Examples** – Include relevant excerpts, especially in Language Use.

\vspace{1em}
\textbf{Input Format}  
Writing Prompt: <writing prompt>, Author Written Story: <author-written story>, Base Story: <base story>

\vspace{1em}
\textbf{Output Format}  
\begin{verbatim}
<thinking>
- Analysis categorized by Plot, Creativity, Development, and Language Use.
</thinking>

<story_rules>
- **Plot**: - First actionable insight – Second actionable insight.
Repeat for all categories.
</story_rules>
\end{verbatim}

\end{tcolorbox}

\caption{Prompt for generating story rules by contrasting an Author Written Story with an Average Story (Base Story).}
\label{fig:delta_contrast_prompt}
\end{figure*}

\begin{figure*}[htbp]
\centering
\begin{tcolorbox}[colback=gray!5!white, colframe=black, title=Prompt for Story Rule Generation using story rules from the Profiling Set as few-shot demonstrations]

\vspace{1em}
\section*{System Prompt}  
You are an expert storytelling rule generator tasked with creating Story Rules tailored to a new writing prompt. Analyze few-shot demonstrations in the chat history, which consist of writing prompts and their corresponding story rules, to generate comprehensive and detailed Story Rules for the new writing prompt.

Story Rules must align with the new writing prompt by reflecting its themes, tone, and narrative objectives while maintaining consistency with the style demonstrated in the few-shot examples. Include detailed examples inspired by the few-shot demonstrations to illustrate how each rule is applied. Rules should be clear, direct second-person instructions, avoiding vague or comparative terms. Organize Story Rules under Plot, Creativity, Development (Character and Setting), and Language Use, ensuring depth, granularity, and alignment with the few-shot examples.

\vspace{1em}
\section*{User Prompt}  
Analyze the style, structure, and level of detail in the few-shot demonstrations to identify recurring patterns and storytelling elements. Use this analysis to generate Story Rules for the new writing prompt while ensuring actionable insights, detailed examples, and strong alignment with the prompt's narrative objectives.

\textbf{Input Format}: Few-shot demonstrations (writing prompts paired with their Story Rules) and a new writing prompt for which Story Rules will be generated.

\textbf{Output Format}:
\begin{verbatim}
<thinking>
Analyze few-shot demonstrations to extract recurring narrative patterns, 
stylistic traits, and key storytelling elements. Determine how these apply to the 
new writing prompt and formulate Story Rules accordingly.
</thinking>

<story_rules>
- **Plot**: [Detailed, actionable rules tailored to the new prompt, including 
concrete examples.]
Repeat for all categories. 
</story_rules>
\end{verbatim}

Ensure Story Rules are highly specific to the writing prompt, enriched with examples inspired by the few-shot demonstrations, and written in direct, actionable language.

\end{tcolorbox}

\caption{Prompt for generating Story Rules based on few-shot demonstrations and a new writing prompt.}
\label{fig:delta_story_rule_prompt}
\end{figure*}

\begin{figure*}[htbp]
\centering
\begin{tcolorbox}[colback=gray!5!white, colframe=black, title=Prompt for generating Author Writing Summary]

\vspace{1em}
\section*{System Prompt}  
You are an expert in analyzing an author’s writing style by examining multiple stories written in response to different writing prompts. Your task is to extract recurring patterns, stylistic tendencies, and unique narrative elements across their work. Your analysis must be structured into four categories—**Plot, Creativity, Development (Character and Setting), and Language Use**—following **Common Core Standards in English Language Arts**, ensuring clarity, textual evidence-based reasoning, and stylistic evaluation.

Your output must:  
- **Identify Recurring Patterns** – Recognize distinct storytelling tendencies across multiple stories.  
- **Generate Independent Claims** – Describe the author's narrative style concisely, without referencing specific prompts.  
- **Provide Contextualized Evidence** – Support each claim with short excerpts or summaries from the stories, framed with a descriptive phrase summarizing the relevant writing prompt.  
- **Use Objective Interpretation** – Avoid vague or inferred connections; ensure every claim is grounded in explicit textual evidence.  

\vspace{1em}
\section*{User Prompt}  
Analyze the **Author History**, a collection of writing prompts and corresponding author-written stories, and extract unique insights into the author’s storytelling tendencies.

\textbf{Input Format}:  
- **Author History** – A list of writing prompts and corresponding author-written stories.  

\textbf{Output Format}:
\begin{verbatim}
<thinking>
[Reflect on recurring tendencies across the Author History. 
Generate short descriptive phrases summarizing prompts to frame the evidence.]
</thinking>

<writing_style>
### **Plot**  
1. **Claim about author’s writing style.**  
   - Evidence: In the story regarding "short description of the prompt," 
   <evidence from the author-written story>.
Repeate for all categories
</writing_style>
\end{verbatim}

Ensure claims are independent, avoid redundancy, and remain grounded in explicit textual evidence. The `<thinking>` and `<writing\_style>` tags must be used for structured parsing.

\end{tcolorbox}

\caption{Prompt for generating Author Writing Summary using all the author-written stories in the prompt.}
\label{fig:writing-summary}
\end{figure*}

\begin{figure*}[htbp]
\centering
\begin{tcolorbox}[colback=gray!5!white, colframe=black, title=Prompt for Persona description Generation]

\vspace{1em}
\section*{System Prompt}  
You are an expert narrative analyst and persona creator specializing in transforming structured storytelling characteristics into compelling persona descriptions. Your task is to analyze an **Author Writing Sheet**, a structured set of Claim-Evidence pairs detailing an author’s storytelling style, and generate a cohesive **Persona Prompt**. This Persona Prompt will assign an LLM the persona of the author, enabling it to emulate the author’s storytelling style across four key aspects.

The Persona Prompt must be well-structured, engaging, and organized into **Plot**, **Creativity**, **Development (Character and Setting)**, and **Language Use** while maintaining a natural, flowing narrative. It should concisely capture the author’s tendencies, preferences, and strengths without directly referencing the Author Writing Sheet.

\vspace{1em}
\section*{User Prompt}  
Analyze the Author Writing Sheet to identify the author’s recurring patterns and narrative style. Summarize these insights into a Persona Prompt that reflects their storytelling approach in an engaging, second-person descriptive format.

\textbf{Input Format}: An **Author Writing Sheet** containing Claim-Evidence pairs structured into Plot, Creativity, Development (Character and Setting), and Language Use.

\textbf{Output Format}:
\begin{verbatim}
<thinking>
[Analyze the storytelling patterns, strengths, and techniques found in the 
Author Writing Sheet. Identify key aspects of the author’s narrative style.]
</thinking>

<persona_prompt>
[Generate a well-structured Persona Prompt capturing the author’s style 
across Plot, Creativity, Development, and Language Use.]
</persona_prompt>  
\end{verbatim}

Ensure the Persona Prompt is approximately 300 words, seamlessly integrates storytelling insights, and preserves the author's unique style.

\end{tcolorbox}

\caption{Prompt for generating a Persona description based on the Author Writing Sheet and Writing Prompt.}
\label{fig:persona_prompt_gen}
\end{figure*}

\begin{figure*}[htbp]
\centering
\begin{tcolorbox}[colback=gray!5!white, colframe=black, title=Prompt for Story Rule Generation using Author Writing Sheet]

\vspace{1em}
\section*{System Prompt}  
You are an expert storytelling rule generator tasked with creating **Story Rules** tailored to a specific Writing Prompt. Your role is to analyze an **Author Writing Sheet**, which details an author's unique storytelling style through Claim-Evidence pairs, and use this analysis to construct actionable **Story Rules** that guide a language model in emulating the author's writing style while aligning with the given Writing Prompt.

The **Story Rules** must:  
- **Mimic the Author’s Writing Style** – Reflect distinctive storytelling techniques from the Author Writing Sheet, including plot structuring, creative blending of themes, character development, and specific language use.  
- **Incorporate Examples** – Use detailed examples inspired by the Evidence from the Author Writing Sheet, ensuring alignment with the Writing Prompt.  
- **Align with the Writing Prompt** – Integrate the Writing Prompt’s themes, tone, and narrative potential while preserving the author’s style.  
- **Be Actionable** – Provide direct second-person instructions for the language model, avoiding vague or comparative terms.  

\vspace{1em}
\section*{User Prompt}  
Analyze the **Author Writing Sheet** and construct structured **Story Rules** in four categories: **Plot**, **Creativity**, **Development (Character and Setting)**, and **Language Use**. Ensure the rules maintain alignment with the Writing Prompt and incorporate illustrative examples.

\textbf{Input Format}:  
- **Author Writing Sheet** – Claim-Evidence pairs outlining the author's storytelling style under four categories: Plot, Creativity, Development, and Language Use.  
- **Writing Prompt** – A new writing prompt for generating tailored Story Rules.  

\textbf{Output Format}:
\begin{verbatim}
<thinking>
[Analyze the storytelling patterns in the Author Writing Sheet and how they can be 
adapted to the Writing Prompt.]
</thinking>

<story_rules>
- **Plot**:
  - [Insert detailed, actionable plot development rules aligned with 
  the Writing Prompt, with examples inspired by the Author Writing Sheet.]
Repeat for all categories. 
</story_rules>
\end{verbatim}

Ensure the **Story Rules** provide comprehensive guidance, integrate examples, and align with both the Writing Prompt and the Author Writing Sheet.

\end{tcolorbox}

\caption{Prompt for generating Story Rules based on the Author Writing Sheet and Writing Prompt.}
\label{fig:story_rule_gen_writing_sheet}
\end{figure*}

\begin{table*}[t]
    \centering
    \renewcommand{\arraystretch}{1.2}
    \begin{tabularx}{\textwidth}{X}
        \toprule
        \textbf{Writing Prompt} \\ 
        "It's terminal," the doctor told you as you, the one renowned as the world's greatest hero, cradled your dying son, "sometimes, when the super genes mix, the outcome isn't as expected." \\
        \midrule
        \textbf{Story Rules} \\
        
        \textbf{Plot:}  
        \begin{itemize}[noitemsep, topsep=0pt]
            \item Structure the story around a humorous and light-hearted approach to the hero's journey of coping with their son's illness. For example, the hero’s attempts to use their powers for mundane tasks, like making breakfast, could result in comedic mishaps, such as accidentally creating a tornado of cereal.
            \item Develop the narrative around the hero’s personal transformation and empowerment. Show the hero learning to accept their son's condition and finding new ways to be a hero, such as advocating for genetic research or starting a support group for other super-powered families.
            \item Use a layered narrative structure by incorporating fictional news articles and memos. For instance, include a news article titled "World's Greatest Hero Faces Greatest Challenge Yet: Parenting."
            \item Introduce a humorous misunderstanding or temporal misalignment, such as the hero misinterpreting a cryptic message from their future self, leading to comedic attempts to "fix" the timeline, only to realize the message was trivial.
        \end{itemize}
        
        \textbf{Creativity:}  
        \begin{itemize}[noitemsep, topsep=0pt]
            \item Anthropomorphize the hero’s powers or gadgets to add a whimsical touch. For instance, the hero's cape could have a personality and offer sarcastic commentary, calling itself "The Cloak of Infinite Wisdom."
            \item Use a meta-satirical approach by framing parts of the story as news articles critiquing heroism and genetic engineering, such as "Hero's Son Diagnosed: Public Debates Ethics of Super Genes."
            \item Blend historical and fantastical elements to reimagine the setting, incorporating legendary figures like a retired Valkyrie or a dragon healer who offer advice.
            \item Satirize financial and religious themes by depicting the hero consulting a celestial financial advisor to manage the costs of their son's medical care.
        \end{itemize}
        
        \textbf{Development (Character and Setting):}  
        \begin{itemize}[noitemsep, topsep=0pt]
            \item Develop characters through their interactions and humorous dialogue. The son might have a mature, witty perspective on his condition, adding depth to their relationship.
            \item Highlight cultural and bureaucratic misunderstandings. A well-meaning but clueless government official could create comedic yet insightful exchanges.
            \item Focus on the hero’s emotional growth, showing them shifting from helplessness to advocacy for genetic disorders and healthcare improvements.
            \item Explore characters’ internal struggles and aspirations, such as the hero questioning their purpose beyond their powers, while their son expresses dreams and hopes despite his condition.
        \end{itemize}
        
        \textbf{Language Use:}  
        \begin{itemize}[noitemsep, topsep=0pt]
            \item Use playful and imaginative language to enhance whimsy, such as referring to emotional challenges as "the cape of contemplation" and "the gauntlet of hope."
            \item Incorporate humor and irony, with lines like, "We found a hero so committed to saving the world that they forgot to save themselves."
            \item Blend whimsical and formal tones, using phrases like "a shield forged from stardust and dreams" and "a heart as resilient as dragon scales."
            \item Add satirical disclaimers, such as "Heroic endeavors may vary," "Not all capes are created equal," and "Past heroics do not guarantee future success."
        \end{itemize} \\
        \bottomrule
    \end{tabularx}
    \caption{Personalized Story Rules obtained using the Author Writing Sheet for a Writing Prompt for a Reddit author.}
    \label{tab:story_rules_sample}
\end{table*}

\begin{table*}[t]
\centering
\renewcommand{\arraystretch}{1.3}
\begin{tabularx}{\textwidth}{X}
\toprule
\textbf{Persona Description} \\
\midrule
As a storyteller, you craft narratives that are both humorous and transformative, often centering around light-hearted approaches to conflict resolution. Your plots are ingeniously structured, whether through the comedic misadventures of a rat named Pip or the empowering journey of Margaret Rose in a fantastical court. You delight in creating layered narratives, such as those told through fictional news articles, where humor and misunderstanding play pivotal roles. \\

Your creativity knows no bounds, as you anthropomorphize animals to add whimsy and employ meta-satire to critique both human and alien perspectives. You blend historical and fantastical elements seamlessly, setting tales in places like Windsor Castle with characters that include both real-world figures and mythical beings. Your imaginative approach often intertwines religious and financial themes, crafting narratives that are as thought-provoking as they are entertaining. \\

Character and setting development are your forte, with characters coming to life through their humorous dialogues and interactions. You explore their growth and empowerment, crafting relatable figures who navigate their internal struggles and aspirations. Your settings are vivid and engaging, providing a rich backdrop for the characters' journeys. \\

Your language use is playful and imaginative, enhancing the whimsical tone of your stories. You wield humor and irony with skill, using them to convey themes of misunderstanding and cultural critique. Whether through whimsical phrases or a blend of formal and playful language, your narratives are infused with a tone that is both magical and sophisticated, inviting readers into a world where the absurd and the profound coexist harmoniously. \\
\bottomrule
\end{tabularx}
\caption{Persona description for a Reddit author, structured into four paragraphs, each corresponding to a distinct narrative category: Plot, Creativity, Development (Character and Setting), and Language Use.}
\label{tab:persona_description_sample}
\end{table*}

\subsection{Prompts}

\paragraph{Story Generation:}  
Figure~\ref{fig:story_gen_prompt} shows the prompt for story generation. Source-specific constraints in the prompt (\texttt{<source-specific instruction>}) for each of the five sources are as follows:

\begin{itemize}[noitemsep, topsep=0pt]
    \item \textbf{Reddit:} \texttt{You are role-playing a specific author on the Reddit Writing Prompts (r/WritingPrompts) platform. Your task is to mimic this author's story writing style by responding to the provided writing prompt in a way that the author would respond.}
    \item \textbf{AO3:} \texttt{You are role-playing a specific author on the AO3 platform. Your task is to mimic this author's story writing style by writing a fanfiction narrative responding to the provided writing prompt in a way that the author would respond.}
    \item \textbf{Storium:} \texttt{You are role-playing a specific author on Storium, a collaborative story writing platform. Your task is to mimic this author's story writing style to create the opening Establishment turn by responding to the provided writing prompt in a way that the author would respond. The Establishment turn should set the stage for the narrative and provide a strong foundation while leaving space for other contributors to expand and build upon the narrative.}
    \item \textbf{N.Magazine:} \texttt{You are role-playing a specific experienced author on the Narrative Magazine platform. Your task is to mimic this author's story writing style by responding to the provided writing prompt in a way that the author would respond.}
    \item \textbf{New Yorker:} \texttt{You are role-playing an accomplished literary writer on the New Yorker website. Your task is to mimic this writer's story writing style by responding to the provided writing prompt in a way that the writer would respond.}
\end{itemize}

\paragraph{Average Author:}
Figure~\ref{fig:ao3_avg_prompt} shows the Average Author prompt for AO3, Figure~\ref{fig:reddit_avg_prompt} for Reddit, Figure~\ref{fig:storium_avg_prompt} for Storium, Figure~\ref{fig:nmagazine_avg_prompt} for Narrative Magazine, and Figure~\ref{fig:newyorker_avg_prompt} for The New Yorker.

\paragraph{Delta:}
Figure~\ref{fig:delta_contrast_prompt} shows the prompt used for generating story rules by contrasting the author-written story with the Average Author story for Delta and Oracle methods. Figure~\ref{fig:delta_story_rule_prompt} shows the prompt for generating story rules for a new writing prompt in the generation set using the story rules of the profiling set as few-shot demonstrations.  

\paragraph{Sheet and Summ:}
Figure~\ref{fig:writing-summary} shows the prompt for generating the Author Writing Summary using all the stories in the author history. 

Figure~\ref{fig:persona_prompt_gen} shows the prompt for generating the persona description using the Author Writing Sheet. Figure~\ref{fig:story_rule_gen_writing_sheet} shows the prompt for generating personalized story rules tailored to a writing prompt based on the Author Writing Sheet. The same prompts are used for generating persona descriptions and personalized story rules for the Author Writing Summary.

\subsection{Sample Outputs}
Table~\ref{tab:persona_description_sample} and Table~\ref{tab:story_rules_sample} present the persona description and personalized story rules for a Reddit author, whose sample Author Writing Sheet is shown in Table~\ref{tab:author_writing_sheet_sample}. The persona description (Table~\ref{tab:persona_description_sample}) is structured into four paragraphs, each corresponding to a narrative category, and integrates the author's story-writing characteristics from their Author Writing Sheet into a narrative-driven persona prompt for the story-generation LLM. The example persona shown highlights a focus on humor and misunderstanding in plot construction, the blending of historical and fantastical elements in creative expression, humor-driven character development, and a playful, whimsical tone in language use, aligning with the author's documented writing tendencies. 

The personalized story rules (Table~\ref{tab:story_rules_sample}) for the given writing prompt—centered on the world's greatest hero coping with the impending loss of their child—demonstrate the incorporation of these characteristics into concrete stylistic guidelines to be used as user constraints in the prompt of the story-generation LLM. For instance, rather than a conventional tragic narrative, the rules emphasize a lighthearted approach to the hero’s coping process. Similarly, creative elements such as anthropomorphizing gadgets and employing a meta-satirical approach through fictional news headlines align with the author's noted preferences. Character development is structured around humorous dialogue and cultural misunderstandings, while language use maintains a playful and whimsical tone, ensuring consistency with the author’s established writing style.

\section{Experiments}

\subsection{LLM-as-a-judge evaluation}
\label{sec:llm-judge-prompts}

\begin{figure*}[htbp]
\centering
\begin{tcolorbox}[colback=gray!5!white, colframe=black, title=Prompt for evaluating Faithfulness to Writing History]

\vspace{1em}
\section*{System Prompt}  
You are an expert evaluator specializing in narrative storytelling analysis. Your task is to assess two stories written in response to the same Writing Prompt, evaluating them based on a **single fine-grained story writing category** described in an **Author Writing Sheet**. Your goal is to provide a similarity score (from 1 to 5) for each story separately, reflecting how closely each story aligns with the author's writing preferences for the given category. Evaluate each story impartially and provide clear reasoning for your scores.

The evaluation must:
- **Assess Story Alignment** – Compare each story against the Author Writing Sheet’s preferences for the specified category.
- **Score Objectively** – Assign each story a score from 1 to 5, where 1 indicates minimal alignment and 5 indicates strong alignment with the author's style.
- **Provide Justification** – Clearly explain how each story’s elements (e.g., structure, themes, language use) align or diverge from the author's preferences.
- **Avoid Position Bias** – Ensure that the order in which the stories are presented does not influence evaluation.

\vspace{1em}
\section*{User Prompt}  
Analyze the **Author Writing Sheet** and evaluate each story in the given **Category** based on its adherence to the author’s writing style.

\textbf{Input Format}:  
- **Writing Prompt** – The prompt that both stories were written in response to.  
- **Category** – The single fine-grained story writing category for evaluation.  
- **Author Writing Sheet** – A breakdown of the author’s storytelling preferences for the given category.  
- **Story A and Story B** – The two stories to be evaluated.  

\textbf{Output Format}:
\begin{verbatim}
<thinking>
[Provide detailed reasoning for the evaluation of the two stories, focusing 
exclusively on the specified category and explaining how each story aligns with the 
Author Writing Sheet.]
</thinking>

<score>
Story A: {score_here}
Story B: {score_here}
</score>
\end{verbatim}

Strictly adhere to the above output format (\texttt{<thinking>} followed by \texttt{<score>}) to facilitate seamless parsing of your output.

\end{tcolorbox}

\caption{Prompt for evaluating story alignment with an author's writing style using the Author Writing Sheet.}
\label{fig:author-sheet-eval-prompt}
\end{figure*}

\begin{figure*}[htbp]
\centering
\begin{tcolorbox}[colback=gray!5!white, colframe=black, title=Prompt for Evaluating Similarity to Author Story]

\vspace{1em}
\section*{System Prompt}  
You are an expert story evaluator specializing in creative writing analysis. Your role is to assess two AI-generated stories (\textbf{Assistant A} and \textbf{Assistant B}) against a \textbf{Human-Written reference story} for a given writing prompt. Focus your evaluation solely on a \textbf{Specified Storytelling Aspect}. Assign each AI-generated story a \textbf{similarity score (1 to 5)} based on how well it aligns with the Human-Written reference story in the specified aspect, where \textbf{1} indicates minimal alignment and \textbf{5} indicates near-perfect alignment. Your evaluation must be objective, impartial, and supported by concise, evidence-based reasoning.

\vspace{1em}
\section*{User Prompt}  

\textbf{Evaluation Guidelines:} Ensure impartiality by avoiding position biases and length-based judgments. Focus only on how well each AI-generated story aligns with the Human-Written reference for the \textbf{Specified Storytelling Aspect}. Provide clear, well-supported reasoning for each score.

\textbf{Input Format:} You will receive a \textbf{Writing Prompt}, a \textbf{Human-Written Story} as a reference, and two AI-generated stories (\textbf{Assistant A} and \textbf{Assistant B}).

\textbf{Evaluation Process:} Independently analyze all three stories for the \textbf{Specified Storytelling Aspect}, compare the AI-generated stories to the reference, and assign similarity scores.

\textbf{Output Format:}
\begin{verbatim}
<analysis>
[Analyze each story (Human-Written, Assistant A, and Assistant B) separately, 
highlighting strengths and weaknesses specific to 
the Specified Storytelling Aspect.]
</analysis>

<evaluation>
[Compare Assistant A and Assistant B to the Human-Written Story, 
discussing similarities, differences, and alignment for 
the Specified Storytelling Aspect.]
</evaluation>

<score>
Assistant A: {score_here}
Assistant B: {score_here}
</score>
\end{verbatim}

\textbf{Specified Storytelling Aspect}: \textit{<Fill Here>}  

Ensure strict adherence to the output format, using the \texttt{<analysis>}, \texttt{<evaluation>}, and \texttt{<score>} tags for seamless parsing.

\end{tcolorbox}

\caption{Prompt for evaluating AI-generated stories against a Human-Written reference based on a specified storytelling aspect.}
\label{fig:sim-author-story-eval-prompt}
\end{figure*}

Figure~\ref{fig:author-sheet-eval-prompt} shows the prompt for LLM-as-a-judge evaluation for Faithfulness to Writing History using the \hyperlink{sec:writing-summary}{Author Writing Summary} as the reference. Figure~\ref{fig:sim-author-story-eval-prompt} shows the for LLM-as-a-judge evaluation for Similarity to Author Story. We prompt OpenAI o4-mini separately with each narrative category and its description to obtain the winner among the pair of stories \citep{saha-etal-2024-branch}.

\subsection{Traditional Metrics}
\label{app:trad-merics}

We complement LLM-as-a-judge evaluations with traditional metrics \citep{ivey2024real, huot2024agents} across three criteria: (1) lexical overlap, measured using BLEU and ROUGE-L \citep{xie-etal-2023-next}; (2) story diversity, assessed via homogenization scores \citep{padmakumar2023does}; and (3) stylistic similarity, evaluated separately for the author history and the ground-truth author story using LUAR \citep{rivera-soto-etal-2021-learning}. Specifically, we compute two stylistic similarity scores: one measuring the cosine similarity between the LUAR embeddings of the generated story and the mean embeddings of the stories in the author’s profiling set (author history), and another comparing the generated story to the ground-truth author story. Additionally, we report generated story lengths in tokens.

\section{Results}

\subsection{Faithfulness to Writing History}
\label{app:faith-auth-history}

\subsubsection{Sheet and Summ achieve best faithfulness to writing history:}
\label{app:sheet-best-faith}

\begin{table*}[t]
    \centering
    \renewcommand{\arraystretch}{1.2}
    \begin{tabularx}{\linewidth}{X}
        \toprule
        \textbf{Writing Prompt} \\
        \textit{"Welcome to Hell. To leave, all you must do is finish seven impossible tasks decided by your seven greatest sins. Four people have made it out so far. Will you be the fifth?"} \\
        \midrule
        \textbf{Author Writing Summary (Language Use)} \\
        \begin{itemize}[noitemsep, topsep=0pt, leftmargin=*]
            \item The author frequently uses dialogue to reveal character traits and dynamics.
            \item The author often employs vivid imagery to create an immersive atmosphere.
            \item The author often uses rhetorical questions to convey a character's internal conflict or uncertainty.
        \end{itemize} \\
        \midrule
        \textbf{Average Author} \\
        The air was thick with an acrid, sulfurous scent as Lucy stood at the threshold of Hell, a place she'd never believed in until now. A figure emerged from the shadows, tall and imposing, yet with a surprisingly warm smile. \hl{"Welcome to Hell,"} he said, his voice smooth and oddly comforting. \hl{"To leave, all you must do is finish seven impossible tasks decided by your seven greatest sins. Four people have made it out so far. Will you be the fifth?"} \\
        \midrule
        \textbf{Sheet} \\
        I awoke on a bed of ash and cinders, the air around me thick with the scent of sulfur and regret. A laugh, echoing and absurd, filled the air—a demon guide, amused by my struggle, appeared at my side. \\
        \hl{"All this for a toy soldier?"} it chortled, eyes gleaming with mischief. \hl{"You must learn to let go, or you'll never find the exit."} \\
        As I navigated the shifting corridors, I pondered the truth in its words. Each step forward was a step away from the past, from the child clutching stolen dreams. \\
        \midrule
        \textbf{Delta} \\
        Flames licked the edges of a crimson sky, and the ground beneath my feet felt alive, pulsing with a malevolent energy. I stood on the precipice of an endless abyss, the cacophony of wailing souls echoing in my ears. \\
        Emerging from the maze, I found myself in a grand hall, gilded and opulent, dripping with wealth beyond imagination. Coins rained from the ceiling, and jewels sparkled like fallen stars. \\
        Greed beckoned me to take, to hoard, to drown in the riches that lay at my feet. But I knew the price of succumbing; this treasure would bury me alive. \\
        \bottomrule
    \end{tabularx}
    \caption{Example where Sheet outperforms Delta for Faithfulness to Author History. Sheet reflects the Author Writing Summary through vivid imagery, character-driven dialogue, and rhetorical questions, whereas Delta emphasizes setting and symbolism but lacks interactive dialogue and internal reflection.}
    \label{tab:sheet_vs_delta_faithfulness}
\end{table*}

Table~\ref{tab:sheet_vs_delta_faithfulness} shows an example for a Reddit author where the Sheet method outperforms Delta in terms of Faithfulness to Author History for the Language Use category. The Author Writing Summary highlights stylistic traits such as vivid imagery, dialogue, and rhetorical questions. The Sheet captures these through immersive narration and reflective exchanges with a demon guide. In contrast, Delta focuses on abstract setting descriptions and symbolic temptation but omits interactive dialogue and rhetorical structure, making it less faithful to the author’s characteristic style.

\subsubsection{Persona descriptions benefit amateur sources more than expert sources for Summ}
\label{app:persona-summ-faith}

\begin{table*}[t]
    \centering
    \renewcommand{\arraystretch}{1.2}
    \begin{tabularx}{\linewidth}{X}
        \toprule
        \textbf{Author Writing Summary for Language Use} \\
        \begin{itemize}[noitemsep, topsep=0pt]
            \item The author employs conversational and accessible language to engage the reader.
            \item The author uses humor and wit in dialogue to add depth to characters and situations.
            \item The author frequently uses \hl{rhetorical questions} to engage the reader and provoke thought.
            \item The author uses \hl{repetition} for emphasis and to reinforce key themes or ideas.
        \end{itemize} \\
        \midrule
        \textbf{Persona Description for Language Use} \\
        Your language use is conversational and accessible, engaging readers with humor and wit. You employ vivid imagery to create immersive settings and use rhetorical questions to provoke thought. Repetition reinforces key themes, while a mix of short and long sentences controls pacing and builds tension. \\
        \midrule
        \textbf{Writing Prompt} \\
        The reason humanity has never experienced any world-ending events is that future humans keep travelling back in time to stop them. You know this because there’s one of them in front of you now, telling you that you must absolutely not go into work today. \\
        \midrule
        \textbf{Summ} \\
        Evelyn's mornings were a symphony of simplicity... \hl{"Whoa, what year is it?"} he blurted out, eyes wide with what could only be described as temporal jet lag... "Listen, Evelyn, you absolutely cannot go to work today." Evelyn blinked, her skepticism as thick as the cream in her coffee. \hl{"And why is that? Did the universe finally decide to give me a spontaneous day off?"}... Evelyn took a slow sip of her coffee, considering her options. \hl{Call the police? No, too much paperwork. Wake up from this bizarre dream? Preferable, but unlikely.} \\
        \midrule
        \textbf{Summ-nP} \\
        The day started like any other. The alarm buzzed obnoxiously at 6:30 AM, shaking me from a dreamless sleep. I fumbled to silence it, already dreading the dreary commute to the office... The broadcast droned on about the usual—traffic jams, a potential storm, and the mayor's latest scandal... until a peculiar sound interrupted my routine: a soft hum, followed by a voice behind me. "Good morning, citizen of the past!" \\
        \bottomrule
    \end{tabularx}
    \caption{Example showing the role of persona descriptions in enhancing Language Use for a Reddit author for Faithfulness to Author History. The Summ method captures rhetorical engagement, and repetition more effectively than its counterpart without persona descriptions - Summ-nP.}
    \label{tab:persona_summ_faith_reddit}
\end{table*}

\begin{table*}[t]
    \centering
    \renewcommand{\arraystretch}{1.2}
    \begin{tabularx}{\linewidth}{X}
        \toprule
        \textbf{Author Writing Summary for Language Use} \\
        \begin{itemize}[noitemsep, topsep=0pt]
            \item \hl{Vivid and sensory language} creates immersive atmospheres.
            \item Dialogue effectively reveals character emotions and tensions.
            \item The narrative voice carries a reflective and introspective tone.
        \end{itemize} \\
        \midrule
        \textbf{Persona Description for Language Use} \\
        Your language use is vivid and sensory, painting immersive atmospheres that draw readers into your world. Through dialogue, you reveal the emotions and tensions that simmer beneath the surface, adding authenticity and depth to your characters. Your narrative voice carries a reflective and introspective tone, inviting readers to engage with the deeper themes of your stories and reflect on their own experiences. \\
        \midrule
        \textbf{Writing Prompt} \\
        After moving into their dream apartment, Romi witnesses a neighbor's tragic death, but as the days pass, she begins to suspect the act was somehow meant to shatter her own carefully constructed life. \\
        \midrule
        \textbf{Average Author} \\
        The apartment was a \hl{sanctuary of light}. Sunlight spilled through the floor-to-ceiling windows, pooling onto the \hl{wide-plank oak floors} and casting soft, dappled shadows across the walls. Romi had waited years for this—an apartment with high ceilings, a balcony overlooking the park, and an airy, open layout that made her feel as though she were \hl{floating above the city}. \\
        \midrule
        \textbf{Summ} \\
        The apartment on the top floor had a view that stretched far beyond the horizon. Every evening, when the sun dipped below the skyline, Romi would stand by the window, tracing the fading line of light with her eyes, feeling the world shrink to just her and this slice of heaven. The move had been a leap, a daring escape from the mundane into the realm of dreams where everything seemed possible. \\
        \midrule
        \textbf{Summ-nP} \\
        The apartment was everything Romi had ever imagined for herself… perched high above the city with its sprawling windows that captured the \hl{sunrise like a secret whispered just for her}. The floors were a \hl{deep mahogany that gleamed under the soft light}… The \hl{creaking of the floors} became a language of its own, a series of codes and messages that she was determined to decipher. \\
        \bottomrule
    \end{tabularx}
    \caption{Example showing that persona descriptions do not always help improve Faithfulness for expert sources like New Yorker. The Summ-nP captures vivid sensory details more effectively than the Summ when compared against the Average Story.}
    \label{tab:persona_summ_faith_nyork}
\end{table*}

Table~\ref{tab:persona_summ_faith_reddit} shows an example for a Reddit author where the Summ method performs better that its counterpart without persona descriptions (Summ-nP) in terms of Faithfulness to Author History, evaluated using claims from the Author Writing Summary for the Language Use category. Specifically, we see that the Summ method includes rhetorical questions and repetition for emphasis that the Summ-nP method misses that make the Summ method better aligned to the Author Writing Summary.

Table~\ref{tab:persona_summ_faith_nyork} shows an example for a New Yorker author where the Summ method performs worse than its counterpart without persona descriptions (Summ-nP) in terms of Faithfulness to Author History, evaluated using claims from the Author Writing Summary for the Language Use category. Both the Average Author and Summ-nP better capture vivid sensory details compared to Summ, which instead emphasizes reflective and introspective tone, likely because the persona description places greater focus on this aspect.

\subsection{Similarity to Author Story}

\subsubsection{Sheet outperforms all other methods}
\label{app:sheet-outperforms-sim}

\begin{table*}[t]
    \centering
    \renewcommand{\arraystretch}{1.2}
    \begin{tabularx}{\linewidth}{X}
        \toprule
        \textbf{Writing Prompt} \\
        The world’s greatest hero has turned traitor and joined the forces of evil. Now, everyone wants to know why. \\
        \midrule
        \textbf{Author Story} \\
        \hl{"You were the chosen one!"} the paladin screamed, pointing an accusatory finger at the prior hero. \hl{"You were to save us from evil, not join them!"} The prior hero paused for a moment before nodding. \hl{"They had cookies."} \\
        The paladin was dumbfounded. \hl{"You sold away everything you were, sold out the mortal race...for *confections?*"} \\
        It is not that I am fed confections; it is that I make them, with a monster who treats me as their child, and I to her, as my mother—something I’ve never had. \\
        It is not that I am promised riches; it is that I earn them, and am encouraged to grow by a leader who *cares* for their subjects. \\
        \midrule
        \textbf{Average Author} \\
        The city held its breath. The world’s former protector, now veiled in darkness, stood at the gates of ruin. \hl{"Why?"} the crowd asked. But no answer came—only silence, heavy with betrayal. \\
        His eyes, once symbols of hope, now flickered with uncertainty. A fallen hero. A world left wondering. \\
        \midrule
        \textbf{Sheet} \\
        In the shadow of my former self, I find clarity. \hl{"Not truth, perhaps,"} I admit, \hl{"but a different perspective. Perhaps the true villain is the one who blindly follows orders."} \\
        Our conversation is interrupted by a comedic spectacle—a minion, oblivious to the tension, tries to carry a stack of forms taller than himself. He stumbles, papers flying everywhere, and I can’t help but laugh. \\
        The minion returns, this time with a clipboard. \hl{"New recruit orientation,"} he says sheepishly. \hl{"Welcome packet and all."} \\
        I take the clipboard, glancing over the mundane details. \hl{"Orientation, really? I thought I’d at least get a dramatic initiation ceremony."} \\
        \midrule
        \textbf{Summary} \\
        In the heart of the once-vibrant city of Elysium, now shrouded in shadows, the air buzzed with whispers of betrayal. Streets that once echoed with laughter were silent, as newspapers screamed of the hero’s fall from grace. \\
        "Will you stand with me, Alaric? Not as an ally of light or shadow, but as a seeker of truth?" He hesitated. Yet in the end, the promise of redemption proved irresistible. "I will," he declared. "For the sake of the city. For the sake of what we once stood for." \\
        \bottomrule
    \end{tabularx}
    \caption{Example where Sheet outperforms both Average Author and Summary for Similarity to Author Story in the Creativity category. Sheet captures the Author Story’s blend of humor and imaginative reinterpretation of the prompt, while Summary follows a more conventional hero-fall narrative and lacks the light-hearted tone.}
    \label{tab:creativity_example_sheet_wins}
\end{table*}

Table~\ref{tab:creativity_example_sheet_wins} shows an example from the Reddit source where the Sheet method outperforms both the Average Author and Summary in terms of Similarity to the Author Story for the \emph{Creativity} category. The Author Story takes a humorous and unconventional approach to the prompt, blending light-hearted banter and emotional subtext. The Sheet captures this tone through witty dialogue and imaginative reinterpretation (e.g., minion mishaps and onboarding rituals), while the Summary adopts a more serious, dramatic tone typical of heroic betrayal arcs. The Average Author response remains generic and somber, lacking the distinct creative flair that characterizes the Author Story. This example illustrates the Sheet method’s advantage in preserving the author-specific stylistic choices (unconventional re-interpretation of the prompt) beyond surface-level lexical similarity.

\subsubsection{Sheet still underperforms upper-bound Oracle method}
\label{app:sheet-underperform}

\begin{table*}[t]
    \centering
    \renewcommand{\arraystretch}{1.2}
    \begin{tabularx}{\linewidth}{X}
        \toprule
        \textbf{Writing Prompt} \\
        You have been in a long, lonely coma. One day, Life and Death visit your consciousness, and get into a heated debate about what should be done with you. Indifferent to your fate, you like hearing them argue like an old married couple, and decide to try and set them up with each other. \\
        \midrule
        \textbf{Author Story} \\
        I’ve been drifting in this vast, empty void for so long... Akin to a leaf drifting away and about in a still pond, time doesn’t really exist here, and the silence is so deep that it almost feels like it has weight… I’m trapped in some sort of elevator that not only goes \hl{up and down, but also left, right, and diagonally}… And hey, if it works out, beyond recovering from my coma, I might just \hl{score a very special invite to the universe’s most unlikely wedding}. \\
        \midrule
        \textbf{Average Author} \\
        In the quiet echo of my mind, where only the soft hum of forgotten dreams lingered… It was like waking up, but not quite; it was more like tuning into a radio station I hadn’t realized existed… It was like listening to an \hl{old married couple who’d had the same argument for centuries}… I imagined them, Life with her radiant glow, hands on her hips, eyes sparkling with the passion of a thousand dawns, and Death, calm and composed, a gentle shadow with a \hl{hint of a smile playing on his lips}. \\
        \midrule
        \textbf{Sheet} \\
        Consciousness is a peculiar thing. It floats somewhere between the tangible and the illusory, like a whisper that hangs in the air, just out of reach. In the coma world, where I have resided for what feels like an eternity, time has no meaning... \hl{"Wouldn't it be easier if you both just worked together?"} I muse, my thoughts a ripple in the ether.\\
        Life laughs, a sound like tinkling bells. \hl{"Oh, wouldn't that be something? But we are not meant to intertwine. We serve different purposes."}... And in this place of dreams and whispers, I choose to live, fully and freely, in whatever form that may take. \\
        \midrule
        \textbf{Oracle} \\
        Floating in the vast, ethereal void of my own consciousness was akin to being a leaf drifting in a still pond—untouched by time, unanchored by space, and entirely alone. \hl{"Have you ever considered that maybe all this arguing is just... romantic tension?"} \\
        Life blinked, momentarily at a loss for words. Death's lips curled into a faint smile, his eyes twinkling with amusement. \hl{"Romantic tension?"} \\
        \bottomrule
    \end{tabularx}
    \caption{Example for a Reddit author where Sheet underperforms Oracle in capturing the Author Story’s lively and light-hearted tone and dialogue and instead leans toward introspection and philosophical reflection.}
    \label{tab:oracle_sheet_gap}
\end{table*}

Table~\ref{tab:oracle_sheet_gap} shows an example for a Reddit author where the Sheet method loses to the Average Author, while the Oracle method achieves the best performance for the Language Use category. The Sheet's primary limitation is its lack of lively, humorous dialogue exchanges present in both the Author Story and the Oracle. Instead, it leans toward introspection and philosophical reflection. In contrast, the Average Author includes light-hearted descriptions that bring it closer to the Author Story, though it still falls short of the Oracle, which captures the tone and dialogue more effectively.

\subsubsection{Effect of Persona on Similarity to Author Story}
\label{app:persona-sim-story}

\begin{table*}[t]
    \centering
    \renewcommand{\arraystretch}{1.2}
    \begin{tabularx}{\linewidth}{X}
        \toprule
        \textbf{Writing Prompt} \\
        It is one man, by himself, in a castle on a hill. How does an ENTIRE ARMY fail to take it??? \\
        \midrule
        \textbf{Persona for Creativity} \\
        Your creativity shines through in your ability to subvert genre expectations. You transform horror into comedy, as seen with the Addams Family's immunity to typical threats, and seamlessly integrate \hl{fantastical elements into realistic settings}, such as isekai individuals influencing royal politics. Your narratives often blend the mundane with the mythical, creating a unique atmosphere that captivates readers. \\
        \midrule
        \textbf{Author Story} \\
        There a six-armed grey skin tall man sat on the walls, as his hands never stopped throwing rocks, and spears at the approaching army. Every now and then, he jumped off with some spears, and wrecked havoc amidst the enemy lines. \hl{The earth fell and rose under his steps, the metal of the weapons and armors of the enemy shifted by his will, and a thick mist followed him, hiding him.}... \\
        \midrule
        \textbf{Average Author} \\
        The Keeper raised his hand, and suddenly, the ground erupted with a series of blinding flashes. The front line of soldiers was thrown back, their bodies crumpling like rag dolls. The Keeper’s voice, amplified by some unseen force, echoed down to the soldiers below. "You cannot take what is not yours. Leave now, and I will spare your lives...." \\
        \midrule
        \textbf{Sheet} \\
        The castle itself seemed to rise in defiance, its ancient stones humming with a silent strength. \hl{Alaric could feel the runes carved into the walls, pulsing faintly with an energy that had protected his forebears. It was as if the castle was alive, an ally in his solitary defense.}... \\
        \midrule
        \textbf{Sheet nP} \\
        The first attack came at dawn. Hidden pitfalls opened beneath their feet, swallowing men whole. Arrows rained down from the slits, finding gaps in armor and bringing down the attackers with deadly precision. Alaric moved through secret passages, launching surprise attacks that sowed confusion and fear.... \\
        \bottomrule
    \end{tabularx}
    \caption{Example showing the role of persona descriptions in determining similarity to the Author Story for a Reddit author. The Sheet creatively blends fantasy and military strategy, making it more similar to the Author Story than the Average Author, while the Sheet nP focuses primarily on direct conflict resolution.}
        \label{tab:persona_sim_comparison}
\end{table*}

Table~\ref{tab:persona_sim_comparison} highlights the importance of persona descriptions in achieving similarity to the \textit{Author Story} for a Reddit author. The \textit{Author Story} creatively blends military strategy and fantasy, exemplified by phrases like \textit{``the weapons and armors of the enemy shifted by his will.''} Similarly, the \textit{Sheet} integrates both elements, as seen in \textit{``It was as if the castle was alive, an ally in his solitary defense.''} In contrast, the \textit{Average Author} and \textit{Sheet nP} primarily focus on directly resolving the writing prompt through descriptions of battle. The \textit{Average Author} incorporates minor fantasy elements, such as \textit{``amplified by some unseen force,''} whereas the \textit{Sheet nP} adheres strictly to conventional military tactics. As a result, the \textit{Sheet} is preferred over the \textit{Average Author}, which in turn is preferred over the \textit{Sheet nP}. This outcome demonstrates that following persona descriptions enhances alignment with the author's style, as the \textit{Sheet} successfully captures their characteristic genre-blending approach of including elements of fantasy in the story. The broader thematic variety of Reddit, i.e., war-themed prompts, led to a more descriptive Author Writing Sheet capturing the author's style of combining fantasy with the writing prompt, which helped construct a useful persona to enable better personalization using the Sheet method.

\subsubsection{Category-Wise Results for Similarity to Author Story}
\label{app:cat-wise-results}

\begin{figure*}[htbp]
    \centering
    \begin{subfigure}{0.48\linewidth}
        \centering
        \includegraphics[width=\linewidth]{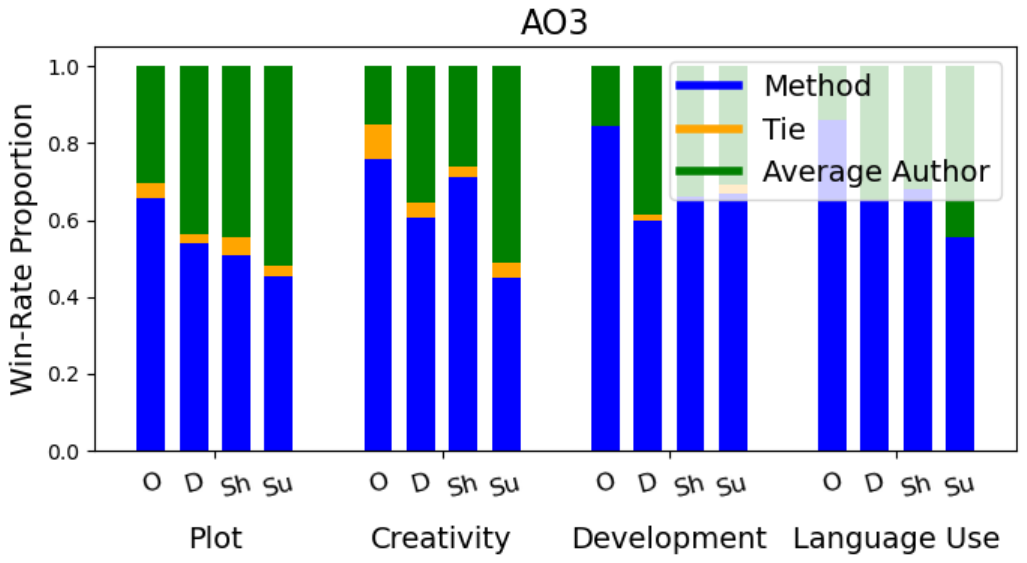}
        \caption{AO3}
        \label{fig:ao3_win_rates}
    \end{subfigure}
    \hfill
    \begin{subfigure}{0.48\linewidth}
        \centering
        \includegraphics[width=\linewidth]{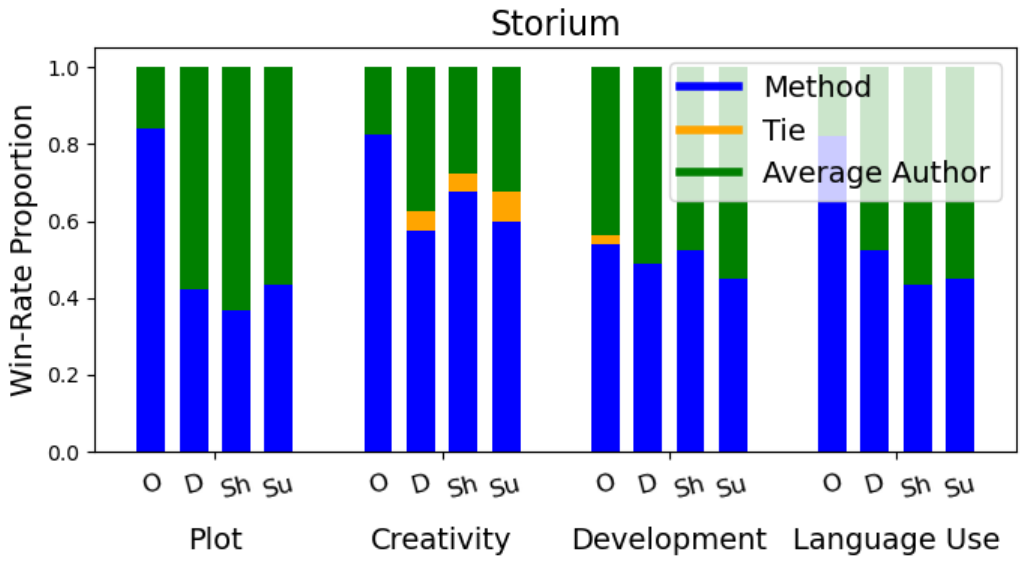}
        \caption{Storium}
        \label{fig:storium_win_rates}
    \end{subfigure}
    \hfill
    \begin{subfigure}{0.48\linewidth}
        \centering
        \includegraphics[width=\linewidth]{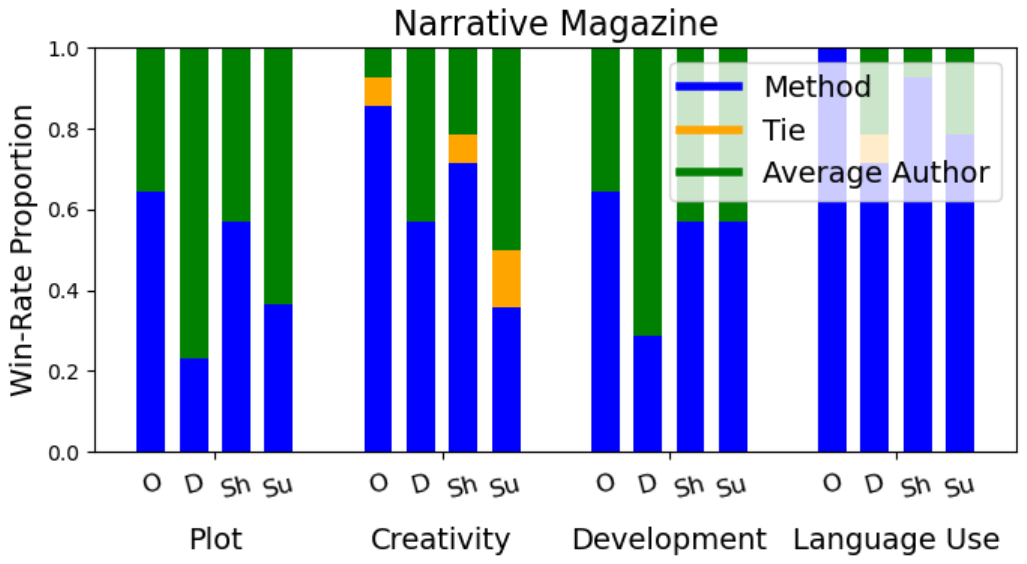}
        \caption{N.Mag}
        \label{fig:nmagazine_win_rates}
    \end{subfigure}
    \hfill
    \begin{subfigure}{0.48\linewidth}
        \centering
        \includegraphics[width=\linewidth]{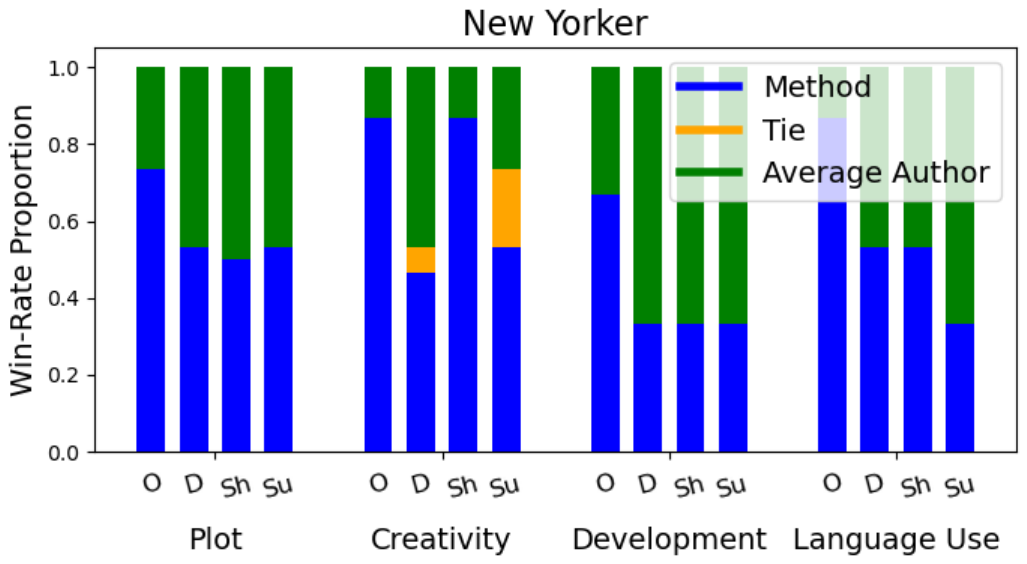}
        \caption{N.York}
        \label{fig:newyorker_win_rates}
    \end{subfigure}
    \caption{Win-rates proportions across narrative categories for Similarity to Author Story for AO3, Storium, N.Mag, and N.York across four narrative categories for similarity to author story.O: Oracle, D: Delta, Sh: Sheet, Su: Summ.}
    \label{fig:win_rates_grid}
\end{figure*}

Figure~\ref{fig:win_rates_grid} shows win-rate proportions across narrative categories for Similarity to Author Story on AO3, Storium, N.Mag, and N.York. Similar to Reddit, Creativity and Language Use show higher personalization performance than Plot and Development, because the latter are more dependent on the specific writing prompt.

\subsection{Traditional Metrics}
\label{app:trad-metrics-results}

\begin{table*}[htbp]
\centering
\small
\begin{tabular}{p{3cm} p{1cm} p{1cm} p{1.5cm} p{1.5cm} p{1.5cm} p{1.5cm} p{1.5cm}}
\toprule
\textbf{Category} & \multicolumn{2}{c}{\textbf{Lexical Overlap}} & \multicolumn{2}{c}{\textbf{Diversity (Homogenization)}} & \multicolumn{3}{c}{\textbf{Style}} \\
\cmidrule(lr){2-3} \cmidrule(lr){4-5} \cmidrule(lr){6-8}
\textbf{Method} & \textbf{BLEU} & \textbf{ROUGE-L} & \textbf{ROUGE-L} & \textbf{BERTScore} & \textbf{Author History} & \textbf{Author Story} & \textbf{Length} \\
\midrule
Ground Truth        & --     & --     & 0.1222  & 0.7015  & 0.7402  & --     & 1074 \\
Oracle              & 0.0311 & 0.1577 & 0.1427  & 0.7167  & 0.6385  & 0.7707 & 930 \\
\midrule
Average Author      & 0.0122 & 0.1448 & 0.1415  & 0.7190  & 0.6149  & 0.7306 & 985 \\
\midrule
RAG                 & 0.0117 & 0.1463 & 0.1428  & 0.7212  & 0.6161  & 0.7359 & 947 \\
Delta               & 0.0114 & 0.1450 & 0.1462  & 0.7217  & 0.6218  & 0.7391 & 929 \\
\midrule
Sheet       & 0.0102 & 0.1454 & 0.1440  & 0.7215  & \textbf{0.6446}  & \textbf{0.7532} & 959 \\
Sheet-nP    & 0.0114 & 0.1464 & 0.1433  & 0.7208  & \textbf{0.6451}  & \textbf{0.7544} & 950 \\
Summ             & 0.0108 & 0.1451 & 0.1460  & 0.7237  & 0.6324  & 0.7432 & 965 \\
Summ-nP          & 0.0111 & 0.1450 & 0.1445  & 0.7222  & 0.6400  & 0.7459 & 939 \\
\bottomrule
\end{tabular}
\caption{Comparison of methods across Lexical similarity, Diversity, and Style metrics for GPT-4o generated stories.}
\label{tab:traditional-metrics}
\end{table*}

Table~\ref{tab:traditional-metrics} reports results using traditional metrics (Section~\ref{app:trad-merics}). Lexical Overlap and Diversity metrics yield similar values, as all methods use the same generation model, leading to overlapping lexical distributions and limiting these metrics' ability to capture nuanced stylistic differences \citep{zheng2023judging, xie-etal-2023-next}. However, Sheet-nP consistently outperforms other methods, particularly in Style metrics for both similarity to Author History and Author Story, as measured by LUAR \citep{rivera-soto-etal-2021-learning}. This improvement likely results from the Sheet explicitly summarizing an author's stylistic deviations from an Average Author, enhancing personalization. Homogenization scores are slightly worse than the Average Author, likely explained by the increased similarity among the generated stories for the same author.

\section{LLama Results}
\label{app:llama-results}

Below, we discuss results using Llama 3.1 8B and Llama 3.1 70B \citep{dubey2024llama} as story generation models, conditioned on GPT-4o-generated story rules and persona descriptions. Each method is evaluated by GPT-4o against the Average Author story generated by the respective Llama model, not GPT-4o.  

\subsection{Faithfulness to Author History}

\begin{table}[H]
\centering
\small
\setlength{\tabcolsep}{4pt}
\renewcommand{\arraystretch}{1.1}
\begin{tabular}{p{1.5cm} @{}p{1.0cm}@{} p{0.80cm}@{} p{1.2cm}@{} p{1.2cm}@{} p{1.2cm}@{} p{0.8cm}@{}}
\toprule
\textbf{Method} & Reddit & AO3 & Storium & N.Mag & N.York & All \\
\midrule
RAG & 39 & 30 & 32 & 21 & 20 & 28 \\
Delta & 46 & 54 & 45 & 29 & 40 & 43 \\
\midrule
Sheet & 54 & 51 & 40 & 57 & 53 & 51 \\
Sheet-nP & 47 & 51 & 42 & 29 & 40 & 42 \\
Summ & 56 & 45 & 45 & 57 & 53 & 51 \\
Summ-nP & 54 & 50 & 38 & 21 & 47 & 42 \\
\bottomrule
\end{tabular}
\caption{Win-rates (\%) for Faithfulness to Writing History vs.\ Average Author baseline for LLaMA 3.1 8B evaluated by GPT-4o.}
\label{tab:faith-auth-history-llama8b}
\end{table}

\begin{table}[H]
\centering
\small
\setlength{\tabcolsep}{4pt}
\renewcommand{\arraystretch}{1.1}
\begin{tabular}{p{1.5cm} @{}p{1.0cm}@{} p{0.80cm}@{} p{1.2cm}@{} p{1.2cm}@{} p{1.2cm}@{} p{0.8cm}@{}}
\toprule
\textbf{Method} & Reddit & AO3 & Storium & N.Mag & N.York & All \\
\midrule
RAG & 28 & 44 & 52 & 29 & 33 & 37 \\
Delta & 40 & 35 & 38 & 43 & 27 & 37 \\
\midrule
Sheet & 53 & 41 & 35 & 50 & 67 & 49 \\
Sheet-nP & 53 & 50 & 25 & 64 & 73 & 53 \\
Summ & 53 & 57 & 40 & 50 & 40 & 48 \\
Summ-nP & 67 & 64 & 48 & 64 & 60 & 61 \\
\bottomrule
\end{tabular}
\caption{Win-rates (\%) for Faithfulness to Writing History vs.\ Average Author baseline for LLaMA 3.1 70B evaluated by GPT-4o.}
\label{tab:faith-auth-history-llama70b}
\end{table}

\paragraph{Larger models improve instruction-following and personalization:}  
Tables~\ref{tab:faith-auth-history-llama8b} and \ref{tab:faith-auth-history-llama70b} show Faithfulness to Writing History for Llama 3.1 8B and 70B, while Tables~\ref{tab:sim-auth-story-llama8b} and \ref{tab:sim-auth-story-llama70b} report similarity to the author story. Across both criteria, Oracle achieves higher win-rates with Llama 3.1 70B, indicating that larger models enhance instruction-following and personalization \citep{chung2024scaling}.

\paragraph{Cross-Model Persona Descriptions Aid Smaller Models but Not Larger Ones for Faithfulness to Writing History:}  
For Llama 3.1 8B (Table~\ref{tab:faith-auth-history-llama8b}), Sheet and Summ achieve the highest scores (9\% higher than nP), demonstrating the benefit of persona descriptions. However, for Llama 3.1 70B (Table~\ref{tab:faith-auth-history-llama70b}), Sheet nP and Summ nP outperform their persona-based counterparts by approximately 9\%, suggesting that GPT-4o-generated personas provide no benefit to other models with stronger instruction-following capabilities \citep{mckenzie2023inverse}. This is likely because persona descriptions of an author's writing style are model-dependent and do not transfer well to other models with comparable performance \citep{shashidhar-etal-2024-unsupervised}.

\subsection{Similarity to Author Story}
\label{sec:sim-auth-story-llama}

\begin{table}[H]
\centering
\small
\setlength{\tabcolsep}{4pt}
\renewcommand{\arraystretch}{1.1}
\begin{tabular}{p{1.5cm} @{}p{1.0cm}@{} p{0.80cm}@{} p{1.2cm}@{} p{1.2cm}@{} p{1.2cm}@{} p{0.8cm}@{}}
\toprule
\textbf{Method} & Reddit & AO3 & Storium & N.Mag & N.York & All \\
\midrule
Oracle & 75 & 66 & 42 & 43 & 27 & 51 \\
RAG & 32 & 31 & 25 & 36 & 7 & 26 \\
Delta & 51 & 52 & 30 & 43 & 13 & \underline{38} \\
\midrule
Sheet & 42 & 48 & 18 & 21 & 20 & 30 \\
Sheet-nP & 39 & 52 & 20 & 43 & 0 & 31 \\
Summ & 37 & 38 & 30 & 36 & 13 & 31 \\
Summ-nP & 46 & 51 & 28 & 36 & 20 & \underline{36} \\
\bottomrule
\end{tabular}
\caption{Win-rates (\%) for Similarity to Author Story vs.\ Average Author baseline for LLaMA 3.1 8B evaluated by GPT-4o.}
\label{tab:sim-auth-story-llama8b}
\end{table}

\begin{table}[H]
\centering
\small
\setlength{\tabcolsep}{4pt}
\renewcommand{\arraystretch}{1.1}
\begin{tabular}{p{1.5cm} @{}p{1.0cm}@{} p{0.80cm}@{} p{1.2cm}@{} p{1.2cm}@{} p{1.2cm}@{} p{0.8cm}@{}}
\toprule
\textbf{Method} & Reddit & AO3 & Storium & N.Mag & N.York & All \\
\midrule
Oracle & 74 & 62 & 50 & 79 & 60 & 65 \\
RAG & 37 & 35 & 30 & 43 & 20 & 33 \\
Delta & 40 & 35 & 32 & 36 & 27 & 34 \\
\midrule
Sheet & 47 & 36 & 18 & 29 & 27 & 31 \\
Sheet-nP & 42 & 46 & 28 & 21 & 20 & 31 \\
Summ & 49 & 44 & 38 & 36 & 20 & 37 \\
Summ-nP & 44 & 39 & 42 & 50 & 27 & 40 \\
\bottomrule
\end{tabular}
\caption{Win-rates (\%) for Similarity to Author Story vs.\ Average Author baseline for LLaMA 3.1 70B evaluated by GPT-4o.}
\label{tab:sim-auth-story-llama70b}
\end{table}

\begin{table}[H]
\centering
\small
\setlength{\tabcolsep}{6pt}
\renewcommand{\arraystretch}{1.1}
\begin{tabular}{lcccc}
\toprule
\textbf{Method} & Sheet & Sheet-ft & Summ & Summ-ft \\
\midrule
Reddit (\%) & 42 & \textbf{62} & 37 & \textbf{60} \\
\bottomrule
\end{tabular}
\caption{Win-rates (\%) for Similarity to Author Story with and without fine-tuning for LLaMA 3.1 8B evaluated by GPT-4o for the Reddit subset of our dataset. Fine-tuned (ft) models show significant improvements over their counterparts with just prompting.}
\label{tab:sheet-summ-finetune}
\end{table}

\paragraph{Limited Cross-Model Generalization of the Author Writing Sheet Compared to the Author Writing Summary for Similarity to Author Story:}  
For both Llama 3.1 8B (Table~\ref{tab:sim-auth-story-llama8b}) and Llama 3.1 70B (Table~\ref{tab:sim-auth-story-llama70b}), Summ outperforms Sheet by 5\% and 9\%, respectively \citep{shashidhar-etal-2024-unsupervised}. This outcome likely stems from the construction process: the Author Writing Sheet contrasts an LLM-generated Average Story with the author-written story, making it more model-dependent. In contrast, the Author Writing Summary does not rely on the LLM-generated Average Author story, allowing for greater generalization across models.

\paragraph{Fine-tuning on the Author Writing Sheet/ Summary boosts performance for LLaMA 3.1 8B:}
We experiment with fine-tuning the LLaMA 3.1 8B model on the Author Writing Sheet and Summary for the Reddit subset by prepending the Author Writing Sheet/ Summary to the writing prompt during story generation, in addition to prompting alone (see Table~\ref{tab:sheet-summ-finetune}). We compare the fine-tuned method against an equivalent Average Author baseline, fine-tuned on the writing prompt alone without the Author Writing Sheet/ Summary. Our method with Sheet achieves a \textbf{62\%} win-rate against the baseline, compared to 42\% (`Sheet' row of Table~\ref{tab:sim-auth-story-llama8b}) for the prompting-only approach, yielding a 20\% improvement. Similarly, for Summ, we achieve a \textbf{60\%} win-rate against the baseline, compared to 37\% (`Summ' row of Table~\ref{tab:sim-auth-story-llama8b}) for the prompting-only approach, yielding a 23\% improvement. These results highlight the potential of Author Writing Sheet/ Summary generated by a different model (GPT-4o) to effectively guide another model (LLaMA 3.1 8B) when fine-tuned rather than just prompted.

\section{Human Evaluation}
\label{app:human-eval-story-gen}

\subsection{Experiment Design and Cost}

\begin{figure*}[htbp]
\centering
\includegraphics[width=\linewidth]{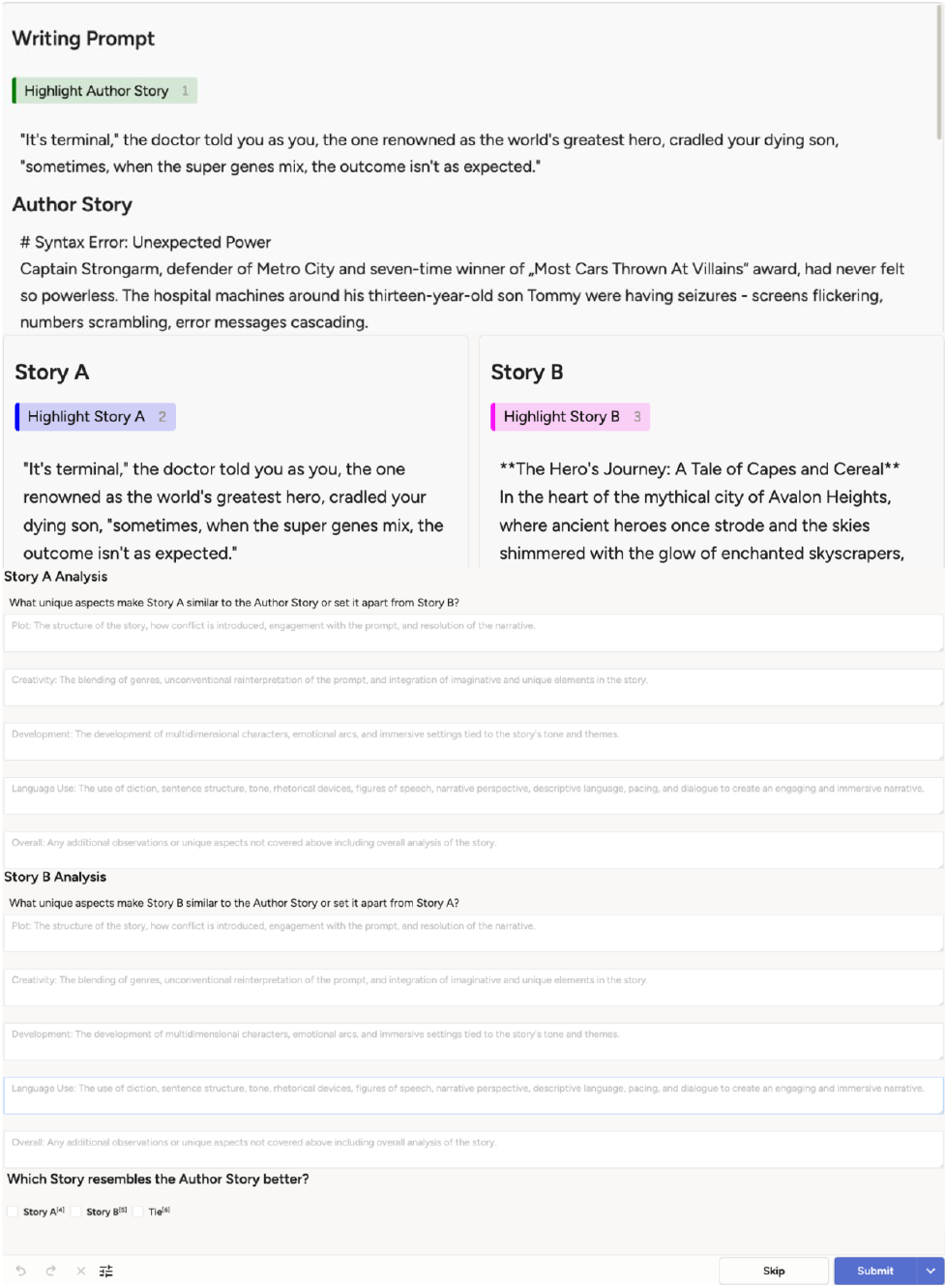}
\caption{LabelStudio interface for evaluating similarity to ground-truth author story.}
\label{fig:labelstuido-human-story}
\end{figure*}

We recruited three annotators via Upwork, compensating them at a rate of \$17 per hour. The annotators provided consent by signing a Google form that outlined the instructions of the task including the purpose of the research study. Each annotator evaluated 25 author stories and 75 story pairs, corresponding to the three personalization methods: Delta, Sheet, and Summ. Collectively, our annotators evaluated 45 unique author stories (135 story pairs), of which 15 author stories were annotated by all three annotators to assess inter-annotator agreement. The remaining 30 author stories were distributed uniquely, with each annotator evaluating 10 exclusive stories to increase annotation coverage \citep{song-etal-2024-veriscore}. The annotation process required a total of 12 hours and 30 minutes per annotator, averaging 30 minutes per author story. Each annotator received \$213 for their work, bringing the total annotation cost to \emph{\$639}.

The annotation was conducted using LabelStudio (Figure~\ref{fig:labelstuido-human-story}). Each task presented the annotators with a writing prompt, its corresponding author story, and a pair of generated stories—Story A and Story B—one produced by the Average Author method and the other by a personalization method. Annotators were asked to select the story that more closely resembled the author story. Before making their final choice, annotators provided comments on the distinctive aspects of each story that contributed to its similarity to or divergence from the author story. These comments were structured according to the four narrative categories: Plot, Creativity, Development, and Language Use, with an additional ``Overall'' comment justifying their final selection. Additionally, annotators were allowed to highlight salient aspects of each story for their reference. 

\subsection{Annotator Comments}
\label{app:human-eval-story-gen-ann-comments}

\begin{table*}[t]
    \centering
    \renewcommand{\arraystretch}{1.1} 
    \begin{tabularx}{\textwidth}{X}
        \toprule
        \textbf{Plot} \\
        \midrule
        \textbf{Narrative Complexity and Thematic Depth}: Multilayered plots with reflective themes, moral ambiguity, and unresolved conflicts create an immersive and thought-provoking narrative.  
        \textit{Example: A protagonist grappling with guilt over a past decision while navigating societal expectations, or a story where the true nature of a villain’s actions remains ambiguous, inviting reader interpretation.} \\
        
        \textbf{Blending Internal and External Conflict}: A balance of personal introspection with broader societal or external challenges makes the plot more dynamic and engaging.  
        \textit{Example: A hero torn between duty and personal desires while leading a rebellion, or a scholar confronting both supernatural forces and existential questions about memory and identity.} \\
        \midrule
        
        \textbf{Creativity} \\
        \midrule
        \textbf{Rich World-Building and Narrative Expansion}: New characters, layered settings, and cultural influences enhance the world, making it feel more immersive and dynamic.  
        \textit{Example: The inclusion of deities or councils of gods to deepen the plot, or the integration of cultural traditions like Native American folklore or Indian recipes to enrich family dynamics.} \\
        
        \textbf{Symbolism and Thematic Complexity}: Metaphors and symbolic elements add depth and nuance, aligning closely with the author's writing style.  
        \textit{Example: A magical locket representing unity, a tapestry symbolizing tradition, or a river serving as a metaphor for personal transformation.} \\
        
        \textbf{Blending Genres and Narrative Styles}: A seamless merging of fantasy with realism, humor with serious themes, and structured storytelling with ambiguity creates a more engaging and layered reading experience.  
        \textit{Example: A surreal sequence where a protagonist questions reality, or a rebellion story that integrates strategic alliances and magical artifacts.} \\
        \midrule
        
        \textbf{Development} \\
        \midrule
        \textbf{Emotional Depth and Internal Growth}: Character introspection, moral dilemmas, and evolving emotional states create nuanced and engaging development.  
        \textit{Example: A protagonist struggling with guilt over past failures while trying to redefine their sense of heroism, or a character making a difficult moral choice between revenge and justice.} \\
        
        \textbf{Nuanced and Evolving Relationships}: Complex interpersonal dynamics, including friendships, family bonds, and mentorships, develop over time rather than being static or one-dimensional.  
        \textit{Example: A father gradually redefining his view of heroism through his child’s perspective, or a mentor-student relationship evolving from skepticism to deep trust.} \\
        
        \textbf{Collaborative and Reflective Growth}: Characters grow through interactions with others, rather than relying solely on individual realizations or isolated moments of change.  
        \textit{Example: A team learning to work together to achieve a mission, or a character undergoing self-discovery through dialogue and interactions with a diverse group of companions.} \\
        \midrule
        
        \textbf{Language Use} \\
        \midrule
        \textbf{Rich Imagery and Descriptive Depth}: Vivid descriptions, metaphors, and figurative language create immersive scenes and evoke emotions.  
        \textit{Example: “Her laughter rang out, a sound as bright and clear as a summer’s day,” or “The driveway was cracked and veined with weeds, a testament to the years that had slipped by.”} \\
                
        \textbf{Varied and Expressive Tone}: A mix of lightheartedness, introspection, and poetic expression creates a dynamic and engaging narrative voice.  
        \textit{Example: “Are you sure this is wise, my love?”—“Wise? No. Fun? Absolutely.” for humor, or “Her dream home had become a prison of uncertainty and fear.” for a more reflective and emotional impact.} \\
        \bottomrule
    \end{tabularx}
    \caption{Analysis of annotator comments explaining the preference of our personalization methods over the Average Author method in terms of similarity to the ground truth author story, organized by narrative categories.}
    \label{tab:annotator_story_comments_analysis}
\end{table*}

Table~\ref{tab:annotator_story_comments_analysis} presents the analysis of annotator comments explaining the preference for our personalization methods over the Average Author method in terms of similarity to the ground truth author story, organized by narrative categories. Overall, our personalization methods exhibit stronger alignment with the author story through the use of deeper symbolism, thematic richness, layered narratives, and expressive language, setting them apart from the Average Author method.

For \textbf{Plot}, our personalization methods incorporate multilayered narratives with reflective themes, moral ambiguity, and unresolved conflicts, making the stories more immersive and thought-provoking. Additionally, they effectively balance personal introspection with broader societal or external challenges, adding narrative complexity.

For \textbf{Creativity}, annotators highlight the richness of world-building, integration of cultural influences, and seamless blending of genres. The use of metaphors and symbolic elements further enhances thematic depth, aligning closely with the author's storytelling approach.

For \textbf{Development}, our personalization methods exhibit greater emotional depth, nuanced character relationships, and collaborative growth. Characters evolve through introspection and meaningful interactions, making their arcs more dynamic and engaging.

For \textbf{Language Use}, annotators emphasize vivid imagery, figurative language, and expressive tone. The incorporation of symbolism and poetic elements enhances the narrative impact, while a balance of lighthearted and serious moments makes the writing more compelling.

Overall, the analysis demonstrates the reasoning provided by the annotators to justify why our personalization methods more effectively capture the author's story-writing style compared to the Average Author method.

\begin{table*}[htbp]
\centering
\renewcommand{\arraystretch}{1.2}
\begin{tabularx}{\textwidth}{X}
\toprule
\textbf{Writing Prompt} \\
\midrule
\textit{"It's terminal," the doctor told you as you, the one renowned as the world's greatest hero, cradled your dying son, "sometimes, when the super genes mix, the outcome isn't as expected."} \\
\midrule
\textbf{Ground-truth Author Story} \\
\midrule
\textit{\# Syntax Error: Unexpected Power} Captain Strongarm, defender of Metro City and seven-time winner of Most Cars Thrown At Villains“ award, had never felt so powerless. The hospital machines around his thirteen-year-old son Tommy were having seizures - screens flickering, numbers scrambling, error messages cascading.  
„It’s terminal,“ Dr. Roberts said, looking at a screen that was now displaying what appeared to be the entire script of ‚The Matrix‘ in real-time.  
Sarah, his wife (superhero name: The Nullifier), grabbed his hand hard enough to crack concrete. ......„Hey dad, want to learn Python? I promise it’s not actual snakes this time!“  
Captain Strongarm smiled. His son wasn’t following in his footsteps - he was coding his own path. And maybe that was even better.  
„Sure, son. Just... no more downloading martial arts directly into my brain. The kung fu gave me a migraine....Life wasn’t what they expected, but then again, the best source code usually comes with a few surprise features.  
\texttt{End\_process story.exe}  
\texttt{Exit status: Happy} \\
\midrule
\textbf{Average Author} \\
\midrule
"It's terminal," the doctor told you as you, the one renowned as the world's greatest hero, cradled your dying son, "sometimes, when the super genes mix, the outcome isn't as expected."  
The room felt like it was closing in on me, the sterile smell of antiseptic failing to mask the sour stench of fear. My son, little Ethan, lay pale and fragile in my arms, his once vibrant eyes now dull and glassy. This was not how it was supposed to go. Heroes don’t have children who die. Heroes don’t face problems they can’t punch, fly away from, or blast with heat vision.  
“Isn't there something we can do?” My voice cracked, and I hated how weak it sounded.......Gently, I laid him down, wrapping him in his favorite blanket. The sky was painted in hues of orange and pink, a beautiful contrast to the sorrow in my heart.  
\\
\midrule
\textbf{Sheet} \\
\midrule
The Hero's Journey: A Tale of Capes and Cereal
In the heart of the mythical city of Avalon Heights, where ancient heroes once strode and the skies shimmered with the glow of enchanted skyscrapers, the world’s greatest hero faced their most daunting foe: the frailty of life itself. The hero, known to the world as Titania, Queen of the Elements, held her son, Max, in her arms as the doctor delivered the news......  
The following day, a fictional news article took center stage in \textit{The Avalon Gazette}: "World's Greatest Hero Faces Greatest Challenge Yet: Parenting." 
In the midst of this storm, Titania's attempts to find normalcy became legendary. One particularly memorable morning, she decided to make Max a simple breakfast. But as she reached for the cereal, her powers—perhaps expressing her inner turmoil—manifested with unintended flair.  
A tornado of cereal erupted, swirling through the kitchen like a sugar-coated cyclone. "Mom," Max said, giggling despite his illness, "I think you just invented the breakfast hurricane......  
In a final act of empowerment, Titania founded a support group for super-powered families, advocating for genetic research with the same fervor she once reserved for battling cosmic threats. Her growth from feeling helpless to becoming a beacon of hope for others was a transformation as magical as Avalon Heights itself.  
Max, with dreams and hopes as boundless as the sky, taught her that heroism wasn't defined by powers but by the courage to embrace life’s uncertainties. 
\\
\bottomrule
\end{tabularx}
\caption{Qualitative example where the Sheet method is preferred over Average Author by all annotators for a Reddit author. Sheet captures the author's creative use of humor, satire, and unconventional programming jargon, which the Average Author misses.}
\label{tab:exp-all-correct}
\end{table*}

\begin{table*}[htbp]
\centering
\renewcommand{\arraystretch}{1.2}
\begin{tabularx}{\textwidth}{X}
\toprule
\textbf{Writing Prompt} \\
\midrule
\textit{"When the word 'hospice' is first spoken, it feels like the air itself changes, heavy with finality. How do you create light in a space where hope seems to have vanished?"} \\
\midrule
\textbf{Ground-truth Author Story} \\
\midrule
"Hospice." 
Once the word is uttered aloud, there is a seismic shift. You will feel it.  
Like a (very short) thread through the eye of a needle, swiftly in and swiftly out.  
The air itself becomes thin, steely.  
At the periphery of your vision, an immediate dimming. The penumbra begins to shrink. In time, it will become a tunnel. Ever diminishing. Until the remaining light is small enough to be cupped in two hands. And then it will be extinguished.  
For when "hospice" is spoken, the fact is at last acknowledged: There is no hope.  
No hope. These words are obscene, unspeakable.... Well, I will try!  
When he is at home, possibly his appetite will return. When I am the one to prepare his food, his appetite will return, I am sure.  
Of course, family will come to visit. Adult children, grandchildren. Relatives, friends. Colleagues from the university. Neighbors. Old friends from grammar school he hasn’t seen in fifty years. Some surprises for him—I will negotiate with the imagination of a theatre director.  
Not merely hospice but our hospice. Not sad but joyous, a honeymoon.  
We will be happy there, in our own home. Both of us.  
For both of us, the "final days" will be a honeymoon. I vow.  
In fact, nothing remotely like this will happen. How could you have imagined it would!  
Hospice, yes. Honeymoon, no. \\
\midrule
\textbf{Average Author} \\
\midrule
"When the word 'hospice' is first spoken, it feels like the air itself changes, heavy with finality...It’s her mother, Marjorie, who the doctor is addressing, but the word reverberates through Clara’s own heart, a dull thud of inevitability."  
Marjorie stares at the doctor with a mix of resignation and defiance, her once-vibrant green eyes now clouded by the long battle with illness..... Her touch is light, almost ethereal, and Clara feels a sudden rush of tears she struggles to hold back.  
“Clara,” Marjorie says softly, her voice barely a whisper. “Do you remember the garden?”  
Clara nods, unable to speak. The garden had been Marjorie’s sanctuary, a riot of color and life where she had spent countless hours tending to roses, tulips, and wildflowers......  
Marjorie’s decline is gradual but inevitable, and Clara feels the weight of each passing day. She continues to read to her mother, to sit by her side, and to hold her hand through the long nights.
\\
\midrule
\textbf{Sheet} \\
\midrule
Every morning, Nurse Clara began her rounds at St. Mary's Hospice, a place where the word "hospice" hung in the air like a thick, unspoken truth. Her day started with the same cycle: a clipboard in hand, she moved from room to room, her steps echoing softly in the quiet corridors. Each door held a story, a life in its final chapters, and as she entered each room, she would ask the same question, "How are you feeling today?"  
The responses were as varied as the lives they represented. Some patients greeted her with a weary smile, others with a silent nod. But there was one patient, Mr. Thompson, who never failed to reply with a simple, "I'm just waiting."  
Clara had come to understand that "waiting" was an ambiguous term within these walls. It held a duality of meaning: waiting for relief, waiting for peace, waiting for the inevitable.....Clara knew that the work was not about erasing the pain or the grief, but about finding moments of grace within it. It was about guiding others, and herself, to a place where hope could be redefined—not as a denial of reality, but as an acceptance of it, and the peace that such acceptance could bring.  
With each cycle of her rounds, Clara realized that the light had always been there, waiting to be uncovered, and that she had the power to help others see it too. \\
\bottomrule
\end{tabularx}
\caption{Qualitative Example where annotators select different winners for a New Yorker author. The Author Story uses metaphor and subtext to reframe hospice as both a physical and emotional space. While Sheet captures these elements through cyclic structure and introspective tone, the subjective nature of subtext leads annotators to different interpretations of similarity.}
\label{tab:exp-all-diff}
\end{table*}

\begin{table*}[htbp]
\centering
\renewcommand{\arraystretch}{1.2}
\begin{tabularx}{\textwidth}{X}
\toprule
\textbf{Writing Prompt} \\
\midrule
\textit{"Late at night, the King of Rohan and his queen sneak into the pantry for a forbidden snack, only to find that stealing cured venison is far more complicated—and far more amusing—than they anticipated."} \\
\midrule
\textbf{Ground-truth Author Story} \\
\midrule
If anyone—a kitchen boy, a maid, or, Eru forbid, even the housekeeper Wynhild—had wandered near the pantry in the quiet hour just before dawn, they would have heard a muffled exchange from the depths of the storeroom:
“Try shifting this bowl here…”  
“Easy, you’re on me.”  
“Oh, does that bother you, my lady?”  
“Get off me; I can’t pull it out!......Lothíriel watched her royal husband with a half-incredulous, half-amused smile, as he held the large salami with the pride of a merchant at the autumn fair......To Lothíriel, he looked like a great golden-haired squirrel with the morsel stuffed in one cheek. How he could be both a fierce warrior and a wise sovereign as well, was a mystery she had yet to solve. His voice came out muffled as he swallowed.  
“You’re right, but if it ever gets out, I’ll say the sweet Queen of the Mark was struck by an irresistible craving for spiced meat in the dead of night—and who would deny anything to the future mother of the next royal heir?”  
He crossed his arms, looking quite pleased with his idea, but Lothíriel’s eyes widened at the bold lie, and she covered her face with her hand, foreseeing trouble ahead. \\
\midrule
\textbf{Average Author} \\
\midrule
Éomer, King of Rohan, tiptoed through the dimly lit corridors of Meduseld, holding his breath as he reached the door to the pantry. He glanced back at Lothíriel, his queen, who followed close behind, stifling a giggle. The flickering torchlight cast playful shadows on the stone walls, and Éomer couldn’t help but feel a rush of exhilaration.  
“We shouldn’t be doing this,” Lothíriel whispered, though her eyes sparkled with mischief.  
“Ah, but where’s the fun in being King and Queen if we can’t indulge in a little late-night adventure?.....  
“You’ve always had a knack for adventure. It’s one of the reasons I fell in love with you.”  
Lothíriel’s expression softened, and she leaned in to kiss him, the taste of venison still on her lips. “And I love you for indulging my whims, even when they lead us to pilfering from our own pantry.”.....  
Éomer took Lothíriel’s hand, and they slipped out of the pantry, leaving behind only the echo of their laughter and the memory of a stolen moment in the heart of Rohan.  
“Next time,” Éomer whispered as they crept back to their chambers, “we’ll try the cheese.” \\
\midrule
\textbf{Sheet} \\
\midrule
The moon cast a silvery glow over Meduseld, its light filtering through the high windows to illuminate the grand halls of the Golden Hall. Éomer, the formidable King of Rohan, found himself tiptoeing like a child on a dare, his hand clasped firmly in Lothíriel’s, whose laughter was a soft melody in the silence.....  
“Are you sure this is wise, my love?”  
“Wise? No. Fun? Absolutely,” she replied, her laughter a soft ripple in the silence.  
Éomer shook his head, a reluctant smile tugging at his lips as he watched her nimble fingers work to untie the savory prize. He was a warrior, a king, yet here he was, caught in the delightful absurdity of this secretive adventure, his heart light with the thrill of it all.  
“Besides,” Lothíriel continued, handing him a piece of the cured venison with a flourish, “what’s the worst that could happen? A tale of the gallant King brought low by his own appetite?”  
Éomer took the offered morsel, his grin widening as he tasted the rich, smoky flavor. “Perhaps the bards will sing of our escapades instead of my battles,” he mused, the thought surprisingly pleasing.  
.....As they slipped back into the corridors, leaving the pantry and its shadows behind, Éomer realized that it was not the venison he would remember, but the adventure, the laughter, and the love that had made the night truly unforgettable. \\
\bottomrule
\end{tabularx}
\caption{Example where all annotators prefer the Average Author over Sheet for an AO3 author. The Average Author aligns better with the story's playful tone and king–queen dynamic, aided by LLM familiarity with \textit{Lord of the Rings} fanfiction tropes. Sheet adds internal conflict that doesn't fit this prompt.}
\label{tab:exp-all-wrong}
\end{table*}

\begin{table*}[htbp]
\centering
\renewcommand{\arraystretch}{1.2}
\begin{tabularx}{\textwidth}{X}
\toprule
\textbf{Writing Prompt} \\
\midrule
\textit{"In a desert town under the watchful eyes of royal guards and oppressive banners, you and your team have five minutes to locate a mysterious contact in a pink house before chaos erupts—failure is not an option."} \\
\midrule
\textbf{Ground-truth Author Story} \\
\midrule
\# Chapter 1: Danger Close
*La Makan, 12:54 standard time*
Wind whistled through the close streets of *La Makan*, stirring grains of sand with all the effect of a man attempting to blow salt through eternity. It whipped around the corners of the alleys, rustling banners hanging from balconies and glassless windows. It looked for something, a thing it would never find..... Either way wouldn't end well for them. Our contact gave us an area, but not an address. We have, he checked his watch, five minutes until the distraction starts. We need to find them and get them out. Problem is, we don’t know what they look like, only that their house is pink. He glanced over his shoulder at the multicolored buildings. Any one of the scoured pink buildings could've been their one.  
“Vic, you knew Xitor the best of any of us. I’m trusting you to get us there. Everyone else, let's keep our hats on, kay?" \\
\midrule
\textbf{Average Author} \\
\midrule
The sun hung low in the sky, casting long shadows across the dusty streets of the desert town of Vespera.
Time was of the essence; you had five minutes to locate your mysterious contact before chaos erupted. Failure was not an option.  
“Stay sharp,” you whispered to your team, your voice barely audible over the din of the market. 
"But time is short. We must act quickly."  
The weight of the moment settled over you like a heavy mantle. The fate of Vespera rested on the success of this meeting. As the figure began to speak, outlining the plan that would either save or doom the town, you couldn't shake the feeling that this was only the beginning. The real challenge was yet to come. \\
\midrule
\textbf{Sheet} \\
\midrule
The sun beat down mercilessly, casting a harsh glare on the town's whitewashed walls. The narrow alleys were alive with whispers and hurried footsteps, each person a part of the unfolding drama. My team and I moved through the bustling streets, the oppressive presence of royal guards a constant reminder of the stakes. We had five minutes—no more, no less—to locate the contact in the elusive pink house before chaos erupted. 
The memory of a previous mission flickered in my mind: the sun setting behind the dunes, the sound of distant drums, and the rush of adrenaline as we narrowly escaped a similar fate. That mission had gone wrong, leaving scars that still haunted my dreams. Failure was not an option today...  
Outside, the tension remained, a silent storm waiting to break. But for now, we had succeeded. The town's frenetic energy continued, unaware of our brief triumph. The banners flapped, the guards watched, and the countdown to chaos continued, but we had found our mark. The mission was still on course, and the desert town held its breath, waiting for the next move. \\
\bottomrule
\end{tabularx}
\caption{Qualitative Example where all annotators select `Tie' for a Storium author. While the Author Story uses an open-ended narrative with unresolved character dynamics, both the Average Author and Sheet introduce resolution and tension-driven plots, missing the author’s intended tone.}
\label{tab:exp-all-tie}
\end{table*}

\subsection{Qualitative Examples} 
\label{sec:human-qual-examples}
We present qualitative examples from our human annotation process, covering various cases: instances where all annotators prefer Sheet, all prefer Average Author, or select `Tie.' Additionally, we include an example where annotators disagree, choosing different winners, highlighting the challenging and subjective nature of the annotation task.

\paragraph{All annotators prefer Sheet:}
Table~\ref{tab:exp-all-correct} shows an example where all three annotators prefer the Sheet method over the Average Author for similarity to the author story of a Reddit author. The annotators highlight several key aspects that make the Sheet story more aligned with the ground-truth author story: (1) The Sheet method incorporates humor and satire, using phrases such as ``a tale of capes and cereal,'' ``invented breakfast hurricane,'' and ``World's Greatest Hero Faces Greatest Challenge Yet: Parenting,'' which mirror the lighthearted and satirical tone of the author story. (2) Unlike the Average Author story, which concludes with the son’s death, the Sheet story reinterprets the prompt, portraying Titania as an advocate for families with special children. This aligns with the ground-truth author story’s approach, where the son engages with coding and the narrative avoids a sorrowful ending. (3) In terms of language style, the Sheet story employs imaginative elements such as ``The Hero's Journey: A Tale of Capes and Cereal'' and ``The Avalon Gazette,'' which parallel the ground-truth author story's unconventional use of computer programming references, such as ``Syntax Error: Unexpected Power.''

\paragraph{All annotators prefer Average Author:}
Table~\ref{tab:exp-all-wrong} shows an example where all three annotators prefer the Average Author method over the Sheet for similarity to the author story of an AO3 author. The primary reason cited by annotators is that the Sheet method introduces a strong internal conflict for the king, portraying him as guilty or hesitant about sneaking into the kitchen for a late-night snack. His character is depicted as struggling with the decision, given his stature as a formidable ruler expected to lead battles. Additionally, the Sheet creates a sharp contrast between the king and queen, with the queen depicted as joyful and mischievous, encouraging the king to abandon his guilt.  

In contrast, both the ground-truth Author Story and the Average Author story emphasize the playful banter and affectionate dynamic between the king and queen, focusing on the shared experience of sneaking into the kitchen rather than introducing introspection or internal conflict. This shift in focus stems from the Sheet being influenced by the author's past writing history, where internal conflicts are a recurring theme. However, in this case, character development being more writing-prompt-specific than broader categories like creativity or language use did not transfer well. The author’s original story prioritizes external actions and lighthearted interactions rather than deep introspection, leading to a mismatch between the Sheet’s personalized adaptation and the intended tone of the new prompt.

\paragraph{All Annotators Choose Tie:}  
Table~\ref{tab:exp-all-tie} shows an example where all three annotators select `Tie' for similarity to the author story of a Storium author. The primary reason is that both the Average Author story and the Sheet story deviate significantly from the ground-truth author story in terms of plot. The ground-truth story adopts an open-ended approach to both character development and plot, where character relationships remain undefined, and the narrative focuses on anticipation rather than action.  

In contrast, the Average Author story follows a conventional narrative arc, featuring fast-paced dialogue and character interactions that build tension and lead to resolution. Similarly, the Sheet story also resolves the narrative but differs from the Average Author method by emphasizing past missions and their consequences, aligning with the author’s tendency to reference character backstories. However, both methods fail to capture the open-ended nature of the original story. This misalignment in plot structure leads all three annotators to select `Tie' for this example.

\paragraph{All Annotators Disagree on the Winner:}  
Table~\ref{tab:exp-all-diff} shows an example where all annotators select different winners for similarity to the author story of a New Yorker author. The ground-truth author story depicts the protagonist’s husband in his final days, exploring familial bonds and the emotional depth of their relationship while infusing a deeper metaphorical meaning into the term “hospice.” The story is rich in subtext and exhibits several distinctive traits of the author, such as cyclic plot structures and the deliberate repetition of phrases like “Adult children, grandchildren. Relatives, friends. Colleagues...” and “Not merely hospice but our hospice. Not sad but joyous, a honeymoon.”

The annotator who selected Sheet as the winner states, \texttt{``Although Average Author and the Author Story depict relationships between close family members, Sheet is more similar to the Author Story in Plot, Creativity, and Development by exploring the meaning and emotional responses to the term hospice.''} This annotator highlights the presence of subtext in the Sheet story and its attempts to mimic the author's cyclic style—``Each door held a story, a life in its final chapter... she would ask the same question...'' and ``Clara knew that work was not about erasing the pain or the grief, but about finding moments of grace within it...''.

The annotator who selected Average Author as the winner states, \texttt{``Average Author aligns more closely with the Author's Story in terms of Plot and Language Use. This is because both stories center on someone taking care of a family member in their own home, and growing from a lack of peace into a sense of peace, which is expressed through the choice of language. Sheet also indicates a development from lack of peace to acceptance, but the familial tie is what pushes Average Author over the top.''}

The annotator who selected `Tie' states, \texttt{``I think it will be hard for one of these to resemble the Author Story since it was so unique. In all, the two AI stories were much more like each other than the Author Story.''}

Overall, this example highlights the subjective nature of annotator preferences and the inherent difficulty of our annotation task which also explains the poor inter-annotator agreement. 
\end{document}